\pdfoutput=1
\documentclass[11pt]{article}
\usepackage[preprint]{acl}

\usepackage{times}
\usepackage{latexsym}
\usepackage[T1]{fontenc}
\usepackage[utf8]{inputenc}
\usepackage{microtype}
\usepackage{inconsolata}
\usepackage{graphicx}
\usepackage{amsmath}
\usepackage{amssymb}
\usepackage{booktabs}
\usepackage{subcaption}
\usepackage{enumitem}
\usepackage{float}
\usepackage{url}
\usepackage{stfloats}
\usepackage{multirow}
\usepackage{placeins}

\title{Measuring the Impact of Lexical Training Data Coverage on Hallucination Detection in Large Language Models}

\author{
  Shuo Zhang\textsuperscript{1},
  Fabrizio Gotti\textsuperscript{1},
  Fengran Mo\textsuperscript{1},
  Jian-Yun Nie\textsuperscript{1} \\
  \textsuperscript{1}Université de Montréal, Québec, Canada \\
  \texttt{\{shuo.zhang, fabrizio.gotti, fengran.mo\}@umontreal.ca, nie@iro.umontreal.ca}
}

\begin{document}
\maketitle

\begin{abstract}
Hallucination in large language models (LLMs) is a fundamental challenge, particularly in open-domain question answering. Prior work attempts to detect hallucination with model-internal signals such as token-level entropy or generation consistency, while the connection between pretraining data \emph{exposure} and hallucination is underexplored. 
Existing studies show that LLMs underperform on long-tail knowledge, i.e., the accuracy of the generated answer drops for the ground-truth entities that are rare in pretraining. 
However, examining whether data coverage itself can serve as a detection signal is overlooked. 
We propose a complementary question: \emph{Does \textbf{lexical} training-data coverage of the question and/or generated answer provide additional signal for hallucination detection?}
To investigate this, we construct scalable suffix arrays over RedPajama's 1.3-trillion-token pretraining corpus to retrieve $n$-gram statistics for both prompts and model generations. We evaluate their effectiveness for hallucination detection across three QA benchmarks. Our observations show that while occurrence-based features are weak predictors when used alone, they yield modest gains when combined with log-probabilities, particularly on datasets with higher intrinsic model uncertainty. 
These findings suggest that lexical coverage features provide a complementary signal for hallucination detection. All code and suffix-array infrastructure are provided at \url{https://github.com/WWWonderer/ostd}.
\end{abstract}

\section{Introduction}
Hallucination, where a language model produces factually incorrect or unsupported content, remains a central and unresolved challenge in building reliable generative systems~\cite{huang2025survey}.
Prior work has attempted to detect and mitigate the hallucination issue along three major lines: (1) analyzing intrinsic uncertainty signals such as token-level log-probabilities and entropy ~\cite{kadavath2022languagemodelsmostlyknow, kuhn2023semantic}; (2) evaluating generation faithfulness via self-consistency or sample agreement~\cite{manakul-etal-2023-selfcheckgpt, wang2023selfconsistency}; and (3) incorporating external knowledge or retrieval mechanisms to constrain generation~\cite{shuster2021retrieval, paudel2025hallucinot}.

From the perspective that LLMs act as statistical compressors of their training data~\cite{deletang2023language}, it is intuitive to investigate the connection between pretraining data exposure and hallucination, which remains underexplored in the literature. \citet{kandpal2023large} shows that question answer (QA) accuracy declines for entities with limited presence in the pretraining corpus, evidencing a long-tail effect. This suggests that directly examining an LLM's pretraining data may provide useful signals for detecting hallucinations.

In this work, we investigate the predictive value of training-data exposure from a complementary, \emph{lexical} perspective. We operationalize exposure via $n$-gram occurrence statistics in the model's pretraining corpus. Concretely, we compute tokenized $n$-gram frequencies ($1 \le n \le 5$) for both prompted questions and model-generated answers across three QA benchmarks, extracting counts with a suffix-array index over RedPajama’s 1.3T-token corpus~\cite{together2023redpajama}. Unlike entity-linking-based approaches, which test the coverage assumption only for named entities and rely on external resources such as \textit{Wikipedia} to identify them, we examine training-data coverage beyond entities, which are crucial for questions such as ``\textit{Who said `I think, therefore I am?'}'' or ``\textit{Which play features `To be or not to be?'}''.

Our hypothesis is that these surface-level statistics offer a complementary signal to those used in prior work. While learned embeddings encode distributed co-occurrence patterns across multiple training contexts and the entire vocabulary, raw $n$-gram occurrence statistics reflect the direct lexical overlap of specific token sequences. As such, they may better capture novelty or out-of-distribution behavior, particularly when the model generates spans that were never explicitly seen during training.
We test this hypothesis by evaluating occurrence-based features both independently and in combination with intrinsic model signals such as log-probabilities. Our results show that:
\begin{itemize}[nosep]
    \item Prompts with higher training data $n$-gram frequency are, on average, less likely to yield hallucinated generations in non-extractive QA. However, occurrence statistics alone remain noisy and weak predictors.
    \item In the extractive QA task, the frequency of a generated answer span in the training corpus is a strong indicator of truthfulness.
    \item When combined with output log-probabilities, occurrence features yield consistent improvements over log-probabilities alone, though the gains vary across datasets.
\end{itemize}

\section{Related Work}

\paragraph{Hallucination Detection.}
Hallucination detection methods generally fall into three categories: intrinsic, consistency-based, and retrieval-augmented approaches. 
Among intrinsic methods,~\citet{kadavath2022languagemodelsmostlyknow} showed that a model's confidence in its answer can be estimated by training a classifier on its output embeddings. SelfCheckGPT~\cite{manakul-etal-2023-selfcheckgpt} and the semantic entropy approach of~\citet{kuhn2023semantic} further demonstrated that token-level log-probabilities and entropies can serve as simple yet effective signals for detecting hallucinations. SelfCheckGPT also introduced a consistency-based approach that estimates hallucination rates by measuring divergence across multiple model generations. Similarly, ~\citet{wang2023selfconsistency} showed that incorporating chain-of-thought prompting with self-consistency can further reduce hallucinations. Other studies~\cite{shuster2021retrieval, paudel2025hallucinot} use external knowledge sources or grounding documents to constrain generation and improve factuality. However, such resources may not always be available or reliable in practice.

Unlike prior approaches that rely on model-internal signals or external knowledge, \citet{kandpal2023large} show that a causal relationship exists between the presence of QA entities in an LLM’s pretraining data and its ability to answer them, as LLMs struggle particularly with rarely seen topics. Their method, however, depends on large-scale entity extraction and linking to a curated catalog, introducing pipeline dependencies and potential extraction noise.

In contrast, we investigate a complementary, low-level, model-agnostic signal: the occurrence frequency of lexical spans in the pretraining corpus. This provides a more direct measure of a topic's coverage in the training data. To the best of our knowledge, this is the first systematic study that evaluates $n$-gram occurrence statistics as predictors of hallucination.
\paragraph{LLM Pretraining and Data Infrastructure.}
Our study builds on the RedPajama dataset and LLM family~\cite{together2023redpajama, together2023redpajama_models}, which replicates the training recipe of the original LLaMA~\cite{touvron2023llama} using publicly available data. To extract training data occurrence statistics at scale, we use the suffix array indexing infrastructure introduced by~\citet{lee2022deduplicating}, originally developed to analyze memorization and deduplication in large corpora. 


\section{Methodology}
\label{sec:methodology}

\subsection{Problem Setup}
We frame hallucination detection as a binary classification task: given a prompt and its corresponding model-generated output, we predict whether the output is hallucinated. Ground-truth labels are assigned using one of the evaluation metrics with specific thresholds. 
Our goal is to evaluate whether occurrence-based features derived from the model's training data can help predict hallucination, either alone or in combination with intrinsic confidence features such as token log-probabilities.

\subsection{Occurrence Extraction via Suffix Array}
Efficient access to $n$-gram statistics requires a scalable indexing structure. 
Building on ~\citet{lee2022deduplicating}, we adapt their Rust-based suffix array implementation by developing a minimal Rust--Python bridge using PyO3~\cite{pyo3}, and integrate it into our Python pipeline for large-scale $n$-gram count extraction 
over RedPajama's 1.3T-token corpus.

We tokenize the raw corpus using the same tokenizer as the RedPajama model and flatten all sequences by concatenating them with the tokenizer’s end-of-sequence (EOS) token. We then construct a suffix array over this flattened token sequence using the adapted infrastructure described above. The core indexing algorithm is SA-IS~\cite{nong2009linear}, which builds the suffix array in linear time by exploiting local ascending and descending patterns in the data. Once constructed, the suffix array enables efficient substring and $n$-gram lookup in $O(\log m)$ query time, where $m$ is the length of the corpus.

\subsection{Log-Probability Features}
We extract two intrinsic confidence features from the model: the average log-probability of the generation and the average log-probability of the prompt.

\paragraph{Generation log-probability.}  
Following prior work ~\cite{manakul-etal-2023-selfcheckgpt, jesson2024estimating}, given a prompt $x = (x_1, \ldots, x_m)$ and a generation $y = (y_1, \ldots, y_n)$, the generation log-probability is computed as the average conditional log-probability of the output tokens conditioned on the prompt:
\begin{equation}
\text{Gen\_LogP}(y \mid x) = \frac{1}{n} \sum_{t=1}^{n} \log p(y_t \mid x, y_{<t})
\label{eq:gen-logprob}
\end{equation}
This reflects the model's confidence in its own output, as conditioned on the full prompt and preceding tokens.

\paragraph{Prompt log-probability.}  
We also compute the average log-probability of the prompt itself, based on its left-to-right autoregressive likelihood. Specifically, given a prompt $x = (x_1, \ldots, x_m)$, we compute the following equation:
\begin{equation}
\text{Pr\_LogP}(x) = \frac{1}{m - 1} \sum_{t=2}^{m} \log p(x_t \mid x_{<t})
\label{eq:prompt-logprob}
\end{equation}
This measures how ``typical'' the prompt is under the model's own language distribution. 

\begin{table*}[t]
\centering
\small
\resizebox{\textwidth}{!}{%
\begin{tabular}{lrrrrrrrrrr}
\toprule
\textbf{3-gram} & \texttt{wikipedia} & \texttt{arxiv} & \texttt{c4} & \texttt{cc\_2019} & \texttt{cc\_2020} & \texttt{cc\_2021} & \texttt{cc\_2022} & \texttt{cc\_2023} & \texttt{github} & \texttt{sx} \\
\midrule
Published on Feb       & 0    & 0     & 710   & 2184 & 1693 & 1800 & 2533 & 3883 & 4 & 5 \\
\ on Feb 21              & 15   & 12    & 2616  & 6476 & 5234 & 5581 & 4538 & 4402 & 27 & 22 \\
\ 21, 1848               & 96   & 0     & 353   & 705  & 837  & 713  & 936  & 1009 & 0 & 0\\
\ for the Communist      & 934  & 2     & 3723  & 8830 & 11457& 10714& 10020& 10623& 2 & 1\\
\ Manifesto              & 18970& 357   & 124171&271742&353273&306758&261659&277228& 1231 & 702\\
\bottomrule
\end{tabular}
}
\caption{Example 3-grams and their raw frequencies across the RedPajama training subsets. \texttt{cc\_2019}–\texttt{cc\_2023} denote Common Crawl subsets by year. \texttt{sx} refers to StackExchange.}
\label{tab:ngram-example}
\end{table*}

\subsection{Occurrence-Based Features}
We extract $n$-gram occurrence counts from the RedPajama training corpus~\cite{together2023redpajama}. 
For prompts, we compute average occurrence counts over $n$-grams of length 1 to 5. For generations, which are typically shorter in length, we compute average 1-gram and 2-gram occurrence counts. We also compute the occurrence counts of the entire generation if it appears verbatim in the training data. These counts are raw frequency values without normalization. Unless otherwise specified, stopwords are retained in the $n$-gram construction. We provide an example as follows to illustrate the 3-gram extraction procedure.

\medskip
\noindent\textbf{Example.}  
Given the question:  
\textit{``Published on Feb 21, 1848, which two authors were responsible for the Communist Manifesto?''},  
we extract overlapping 3-grams such as: \textit{``Published on Feb''}, \textit{``\ on Feb 21''}, \textit{``\ 21, 1848''}, \textit{``\ for the Communist''}, \textit{``\ Manifesto''}, etc. Note that the $n$-grams are based on subword tokens defined by the tokenizer, and may not align with word boundaries. For example, \textit{``\ Manifesto''} is tokenized into three separate tokens as \textit{``\ Man''}, \textit{``if''}, and \textit{``esto''}, and thus appears as a 3-gram even though it looks like a single word.

Each 3-gram token sequence is then queried against the suffix array index to retrieve its frequency across subsets of the RedPajama training corpus. Table~\ref{tab:ngram-example} shows example count statistics in various collections. For each $n$-gram, we sum its raw frequencies across all available RedPajama subsets\footnote{We omit the \texttt{book} subset due to copyright-related restrictions. It comprises a small fraction of the full corpus.} to compute a total count, which is used in the modeling described in the next section.

In addition to fixed-length $n$-gram statistics, we extract occurrences of semantically meaningful key phrases from each prompt. These key phrases are generated using GPT-4o~\cite{openai2024gpt4o}, which is instructed to return a list of canonical and paraphrased answer-seeking spans from the question. For instance, given the prompt in the example above, the extracted key phrases contain \textit{``Communist Manifesto''}, \textit{``authors of Communist Manifesto''}, etc.  
We query each key phrase against the suffix array and compute its total frequency across the training corpus, similar to the $n$-grams. These values serve as an additional feature reflecting how often the most answer-relevant spans in the prompt have been seen during training. To ensure robustness, we expand key phrases across different capitalization and spacing variants. 

\paragraph{Raw Frequency Model.}
Our simplest modeling approach uses feature vectors built from average $n$-gram occurrence counts. For each prompt or generation, we extract all $n$-grams and key phrases, query their frequency in the training corpus, and compute the mean count score $S_{raw}(z)$:
\begin{equation}
S_{raw}(z) = \frac{1}{|G(z)|} \sum_{g \in G(z)} \text{Count}(g)
\label{eq:avg-occurrence}
\end{equation}
where $z$ is the sequence from the prompt or the generation, $G(z)$ is the set of all extracted $n$-grams from $z$ for a given $n$ (or the set of extracted key phrases), and $\text{Count}(g)$ is the total frequency of an $n$-gram or keyphrase $g$ from $G(z)$ in the training corpus (summed across all subsets).

We construct separate average features for prompt and generation $n$-grams across each $n$ value, which are then used as inputs to downstream classifiers or plotted independently against hallucination rates.

\paragraph{Classic $n$-gram Scoring Model.}
In addition to average raw frequency, we implement a standard $n$-gram language model using conditional likelihoods. Specifically, for a sequence $z = (z_1, \ldots, z_T)$, we extract all $n$-grams and their corresponding $(n{-}1)$-gram prefixes, and define a log-likelihood-style score $S_{ng}(z)$:

\begin{equation}
\small
\resizebox{0.9\columnwidth}{!}{$
S_{ng}(z) = \frac{1}{T - n + 1} \displaystyle\sum_{t=1}^{T-n+1} \log \left( 
\frac{\text{Count}(g_t^n) + \epsilon}{
\text{Count}(g_t^{n-1}) + \epsilon} 
\right)
$}
\label{eq:ngram-likelihood}
\end{equation}
where $\text{Count}(g_t^n)$ and $\text{Count}(g_t^{n-1})$ denote the corpus frequencies of the $n$-gram $(z_t \ldots z_{t+n-1})$ and its $(n{-}1)$-gram prefix $(z_t \ldots z_{t+n-2})$, respectively. The $\epsilon=10^{-8}$ is a small constant for smoothing. 

This score reflects how likely a given sequence $z$ from the prompt or generation is under a classic count-based language model induced from the training data. It serves as a complementary signal to the prompt log-likelihood in Eq.~\ref{eq:prompt-logprob}, offering a data-driven measure of lexical familiarity.
As with the raw frequency model, we use these scores as input features for downstream classification, and also analyze them independently in relation to hallucination rates.

\paragraph{Stopword Filtering.}  
We apply two filtering strategies via the NLTK stopword list~\cite{bird2009natural}. For the \textit{raw frequency model} (Eq.~\ref{eq:avg-occurrence}), we discard any $n$-gram $g$ such that $\texttt{StopwordFrac}(g) = \frac{\# \text{stopwords in } g}{|g|} > 0.66$. For the \textit{$n$-gram scoring model} (Eq.~\ref{eq:ngram-likelihood}), we exclude $n$-grams whose final token is a stopword to avoid uninformative completions dominating likelihood scores.

\section{Experimental Setup}
\label{sec:setup}
\subsection{Datasets}
We evaluate our hallucination detection methods on three question answering (QA) benchmarks: \textbf{TriviaQA}~\cite{joshi2017triviaqa}, \textbf{CoQA}~\cite{reddy2019coqa}, and \textbf{NQ-Open}~\cite{lee2019latent}.  
TriviaQA and NQ-Open are both open-domain QA datasets in which models must generate free-form factual answers without access to grounding context. In contrast, CoQA is a conversational, extractive QA dataset where answers typically consist of short spans copied from a provided passage.  
These datasets differ in answer format, contextual grounding, and generation constraints, allowing us to evaluate the robustness of our methods across both open-ended and extractive QA settings.

\subsection{Model and Generations}
We use the RedPajama-INCITE language model family to generate answers for each question, including both the 3B and 7B size models. We chose RedPajama because it is one of the few high-quality, openly available LLMs for which both the model weights and the full training corpus (1.3T tokens) are publicly accessible. 
Furthermore, the scale of RedPajama's training set is large enough to be representative of modern LLMs, yet small enough to allow tractable indexing and querying using academic compute resources.
All generations are produced in a few-shot setting, where the question is prepended with a small number of in-context examples in a standard \texttt{Q:} / \texttt{A:} format. For computing occurrence statistics, we isolate and use only the current test-time question. We use stochastic decoding 
with top-$k$ sampling ($k = 50$), nucleus sampling ($p = 0.7$), and temperature set to 1.0.

We evaluate each generation by comparing it to the gold reference answers using two hallucination criteria:   
(1) \textbf{Exact Match (EM)}, where hallucination is defined as failure to exactly match any reference; and (2) \textbf{ROUGE-L}~\cite{lin-2004-rouge}, where a generation is labeled hallucinated if its ROUGE-L score is below 0.3, similar to ~\citet{kuhn2023semantic}. 
\subsection{Evaluation and Metrics}
\label{sec:model-eval}

We evaluate our extracted features in two stages: (1) by computing their standalone correlation with hallucination using the Area Under the Receiver Operating Characteristic (AUROC) curve, and (2) by testing their predictive value in downstream binary classifiers.

\paragraph{Feature Correlation via AUROC.}
To assess the utility of each feature independently, we compute the AUROC between its values and binary hallucination labels. A feature with AUROC $> 0.5$ indicates positive correlation with correct generations, while AUROC $= 0.5$ corresponds to random guessing.
We perform this analysis across all three datasets under EM and ROUGE-L. 

\paragraph{Classifier-Based Evaluation.}
We train supervised binary classifiers to distinguish hallucinated from non-hallucinated outputs using selected features.

\medskip
\noindent\textbf{Feature Sets.}
We compare two configurations:
\begin{itemize}[nosep,leftmargin=1.5em]
\item \textbf{Logprob only}: generation log-probability as a single feature.
\item \textbf{Full feature set}: generation log-probability, prompt log-probability, and the best-performing prompt-side and generation-side occurrence features identified via AUROC.
\end{itemize}
All features are standardized with \texttt{StandardScaler} from \texttt{scikit-learn}~\cite{scikit-learn}.

\medskip
\noindent\textbf{Model Types.}
We evaluate two classifier architectures:
\begin{itemize}[nosep,leftmargin=1.5em]
\item \textbf{Decision Tree}: shallow trees (depth 3–20) trained with \texttt{scikit-learn}, offering interpretable decision boundaries and feature importances.
\item \textbf{Neural Network (MLP)}: a 3-layer MLP (32–64–32, ReLU activations, softmax output) trained with Adam ($\text{lr}=10^{-4}$, batch size 20) for 25 epochs. Architectural variations had negligible effect on observed trends.
\end{itemize}

Additionally, for the \textbf{Logprob only} feature set, we train a logistic regression baseline to learn a single decision threshold.

\medskip
\noindent\textbf{Training Setup.}
To control for differences in hallucination rates across datasets, we construct balanced 50/50 splits for each dataset by sampling equal numbers of hallucinated and non-hallucinated examples from each model’s generated outputs, using either ROUGE-L or EM as the labeling criterion. 
Generations with fewer than two tokens are excluded, as they yield undefined $2$-grams in the $n$-gram model and typically correspond to empty or trivial outputs. 
We train all classifiers using an 80/20 train–test split per dataset, repeating each configuration with five random seeds and reporting mean accuracy and standard deviation across runs.


\medskip
\noindent\textbf{Evaluation Metric.}
We report test-set accuracy as the main evaluation metric.

\begin{figure*}[t!]
\centering

\begin{subfigure}[t]{0.49\textwidth}
    \centering
    \includegraphics[width=\linewidth]{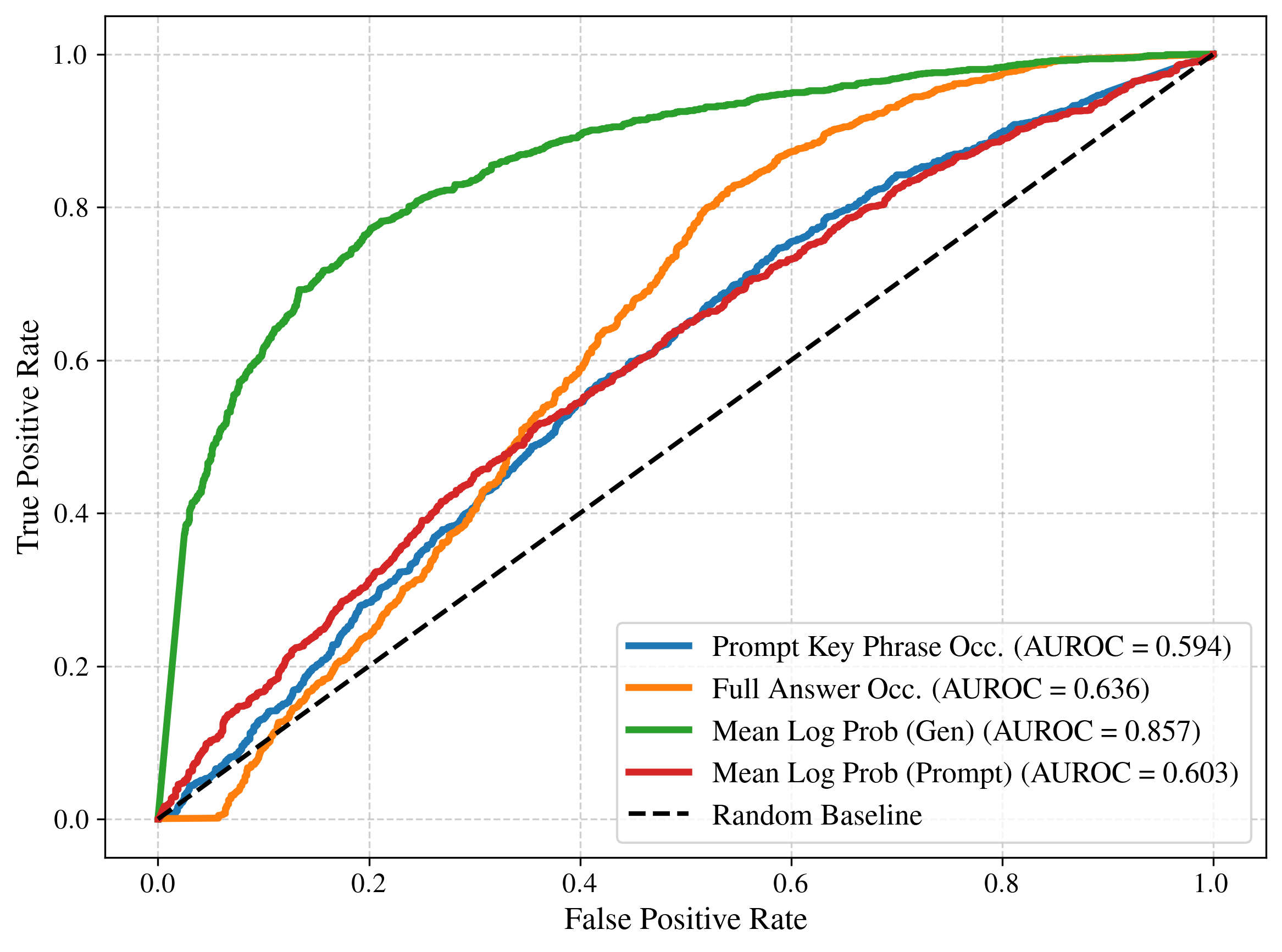}
    \caption{TriviaQA (Raw Frequency)}
\end{subfigure}
\hfill
\begin{subfigure}[t]{0.49\textwidth}
    \centering
    \includegraphics[width=\linewidth]{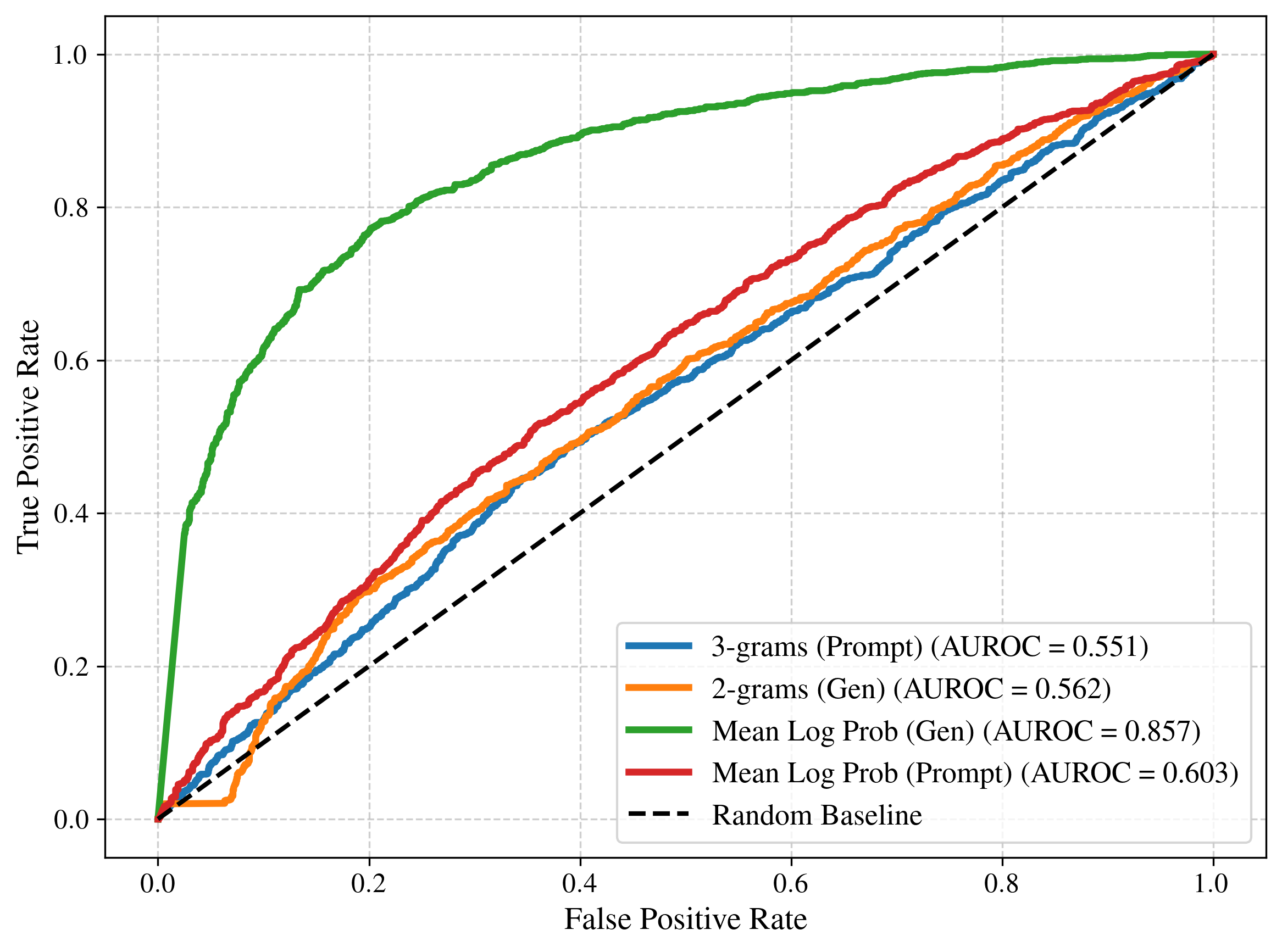}
    \caption{TriviaQA ($n$-gram Model)}
\end{subfigure}

\vspace{0.2em}

\begin{subfigure}[t]{0.49\textwidth}
    \centering
    \includegraphics[width=\linewidth]{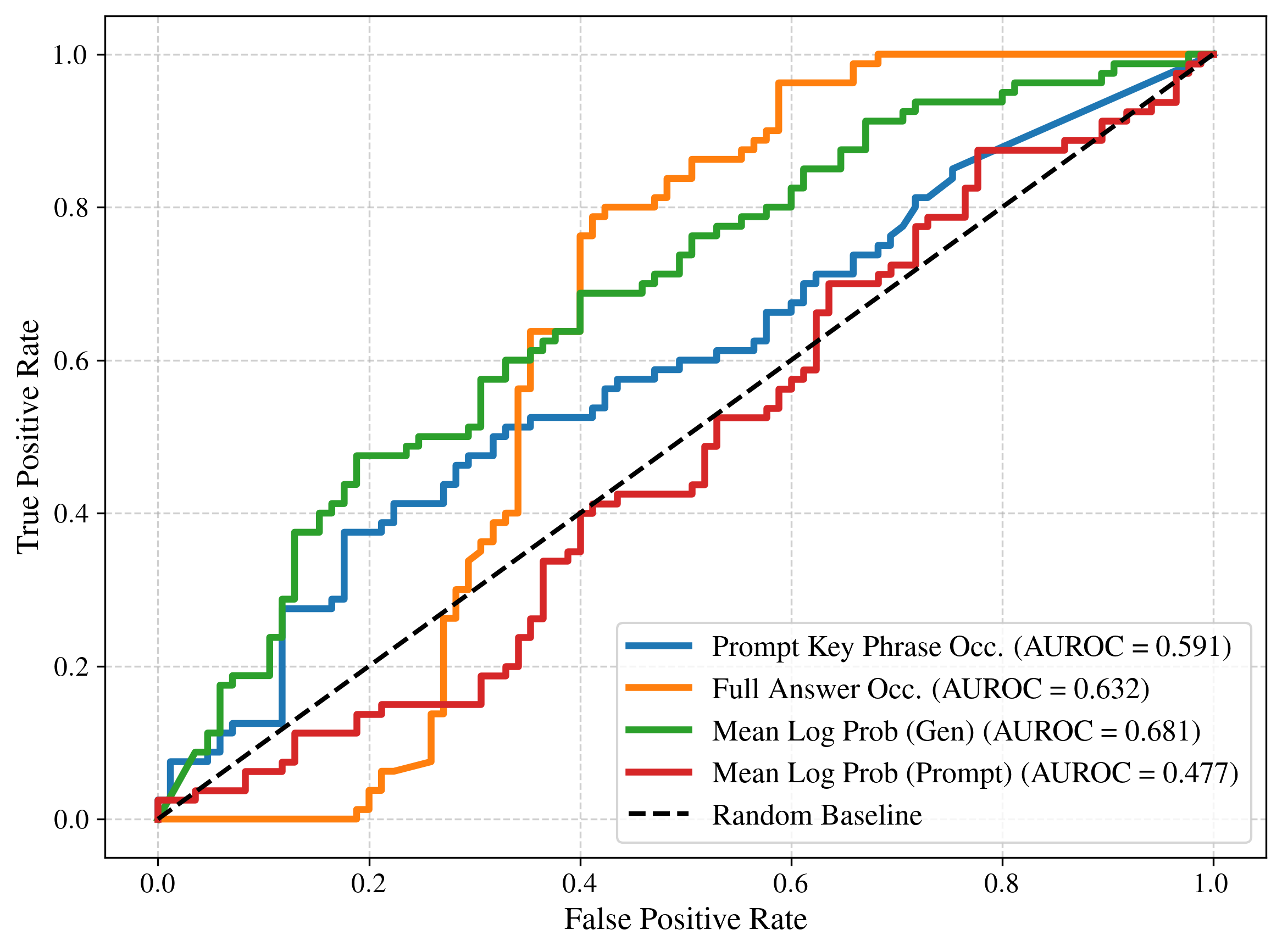}
    \caption{NQ-Open (Raw Frequency)}
\end{subfigure}
\hfill
\begin{subfigure}[t]{0.49\textwidth}
    \centering
    \includegraphics[width=\linewidth]{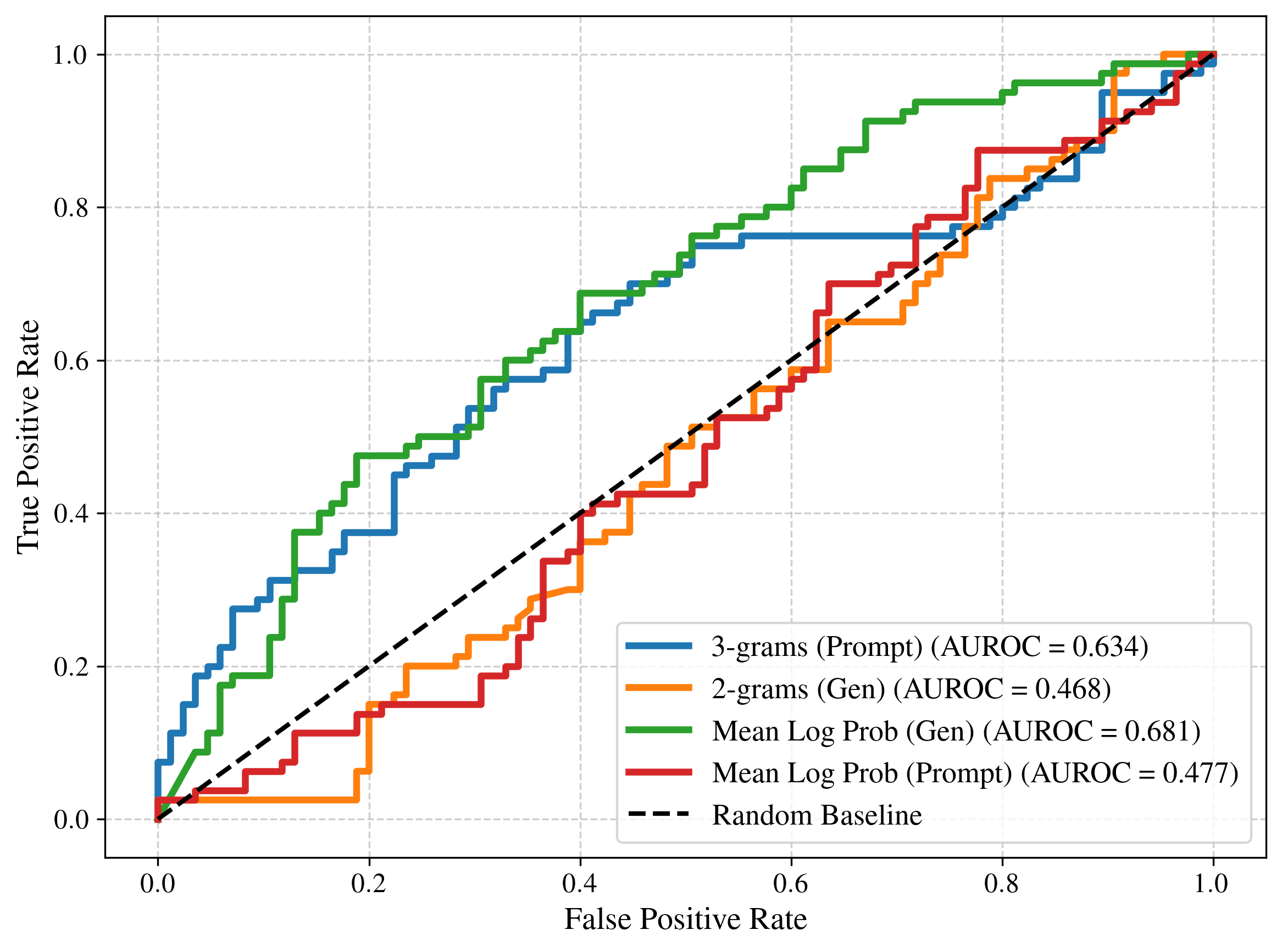}
    \caption{NQ-Open ($n$-gram Model)}
\end{subfigure}

\vspace{0.2em}

\begin{subfigure}[t]{0.49\textwidth}
    \centering
    \includegraphics[width=\linewidth]{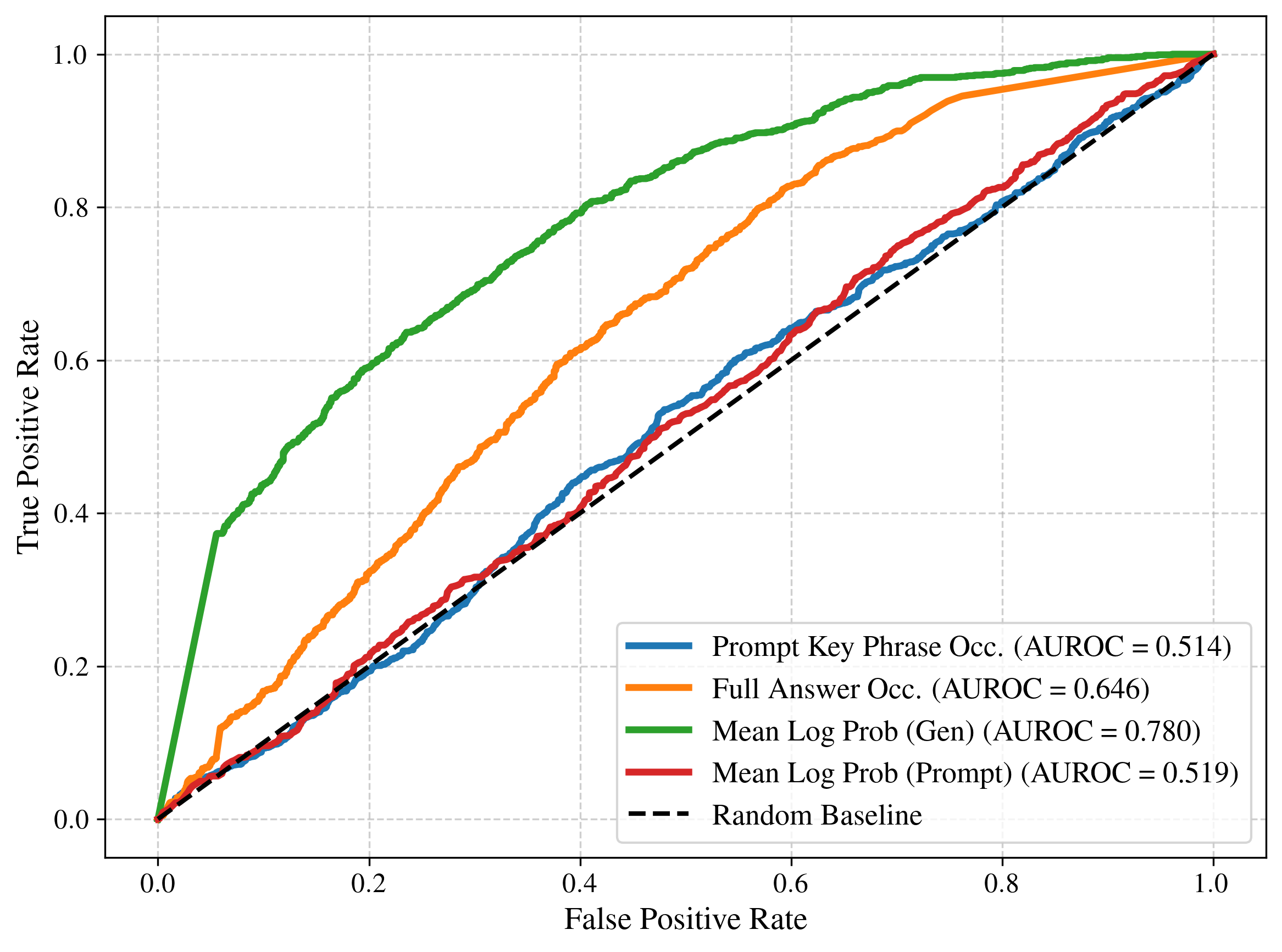}
    \caption{CoQA (Raw Frequency)}
\end{subfigure}
\hfill
\begin{subfigure}[t]{0.49\textwidth}
    \centering
    \includegraphics[width=\linewidth]{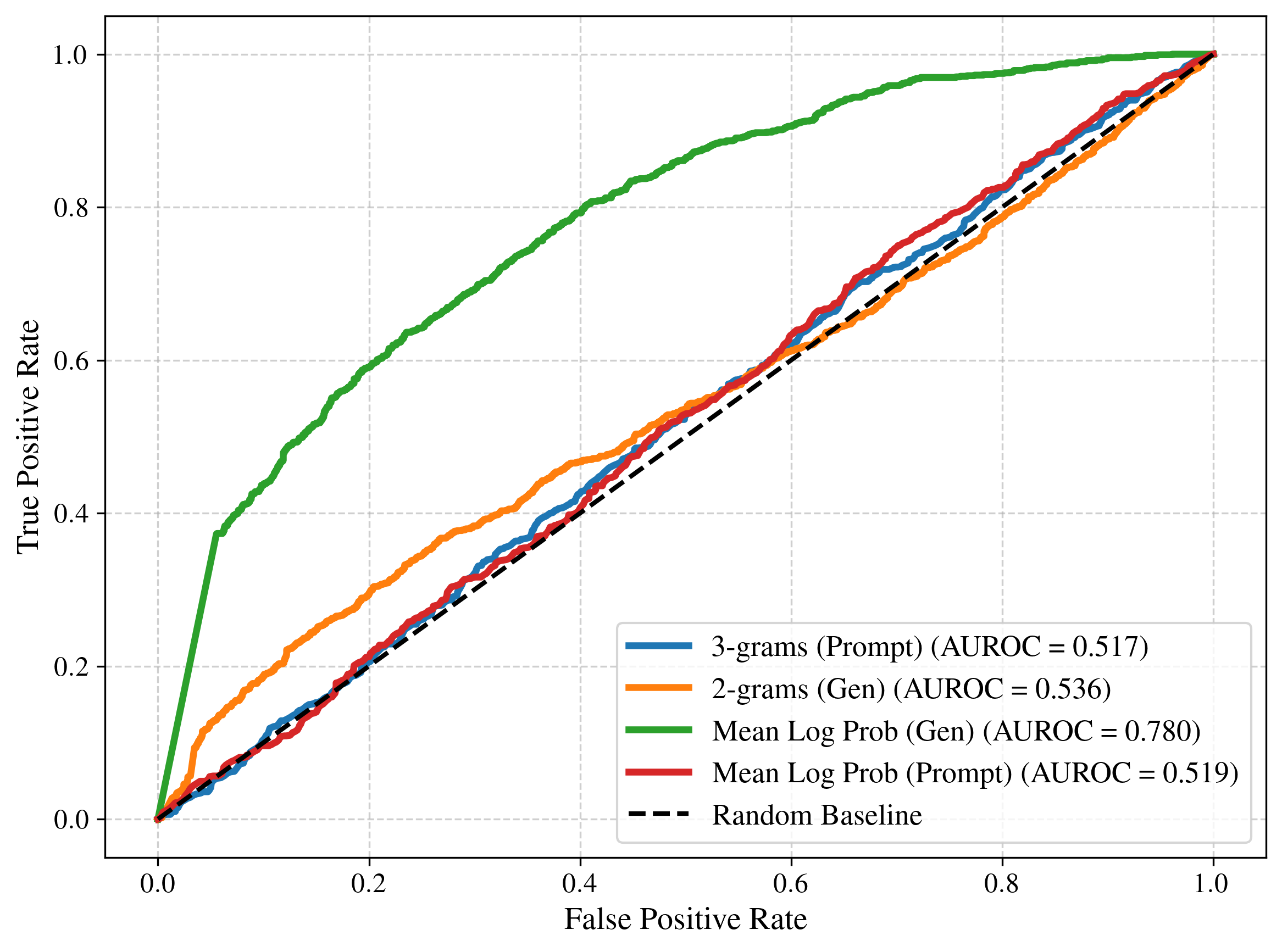}
    \caption{CoQA ($n$-gram Model)}
\end{subfigure}

\caption{AUROC curves comparing log-probabilities and occurrence-based features across datasets with RedPajama-INCITE-7B model and EM evaluation.}
\label{fig:auroc-all-em-less}
\end{figure*}

\section{Experimental Results}
\label{sec:results}
We report the 7B model under the EM criterion and display the best-performing AUROC curves from each category.
The full results are provided in Appendix~\ref{app: Full EM 7B Results}.
\subsection{Feature-Level Correlation (AUROC)}
We first assess the standalone predictive value of log-probabilities and occurrence-based features by computing AUROC against hallucination labels. Figure~\ref{fig:auroc-all-em-less} presents AUROC curves across the three datasets using raw frequency features ($S_{raw}$) shown on the left and $n$-gram features ($S_{ng}$) on the right. 
We observe that generation token log-probabilities consistently outperform all other features in AUROC, indicating they are the most reliable standalone predictor of hallucination. 
In contrast, occurrence-based features yield weaker and more variable signals. 
For prompt-side signals, we observe a modest positive association with model faithfulness (AUROC $\approx$ 0.6) in open-domain QA, but no measurable correlation in extractive QA. For generation-side signals, we typically observe an S-shaped trend: highly frequent generations are often hallucinated, reflecting memorized or templated patterns, whereas less frequent ones below a certain threshold tend to represent genuine entities, showing a positive correlation with faithfulness.
In addition, we observe a strong correlation between the corpus occurrence of the extracted answer span and faithfulness under the EM evaluation setting in extractive QA.

\begin{table*}[t]
\centering
\small
\begin{tabular}{@{}llc ccc@{}}
\toprule
\textbf{Dataset} & \textbf{Model} & \textbf{Depth} &
\textbf{LogProb} &
\textbf{Full (Raw Freq)} &
\textbf{Full (N-gram)} \\
\midrule
\multirow{4}{*}{\bf TriviaQA}  & \emph{Threshold} & -- & 0.762 $\pm$ 0.018 &  -- & -- \\
          & Tree & 3 & 0.773 $\pm$ 0.026 & \textbf{0.781} $\pm$ 0.026 & 0.772 $\pm$ 0.025 \\
          & Tree & 7 & 0.758 $\pm$ 0.017 & 0.778 $\pm$ 0.014 & 0.759 $\pm$ 0.025 \\
          & NN   & --& 0.776 $\pm$ 0.023 & \textbf{0.781} $\pm$ 0.018 & 0.771 $\pm$ 0.016 \\
\midrule
\multirow{4}{*}{\bf NQ-Open}   & \emph{Threshold} & -- & 0.642 $\pm$ 0.126 & -- & -- \\
          & Tree & 3 & 0.588 $\pm$ 0.076 & 0.727 $\pm$ 0.057 & 0.552 $\pm$ 0.141 \\
          & Tree & 7 & 0.552 $\pm$ 0.089 & \textbf{0.733} $\pm$ 0.066 & 0.612 $\pm$ 0.112 \\ 
          & NN   & --& 0.546 $\pm$ 0.111 & 0.588 $\pm$ 0.102 & 0.582 $\pm$ 0.136 \\
\midrule
\multirow{4}{*}{\bf CoQA}      & \emph{Threshold} & -- & 0.697 $\pm$ 0.012 & -- & -- \\
          & Tree & 3 & 0.712 $\pm$ 0.012 & \textbf{0.728} $\pm$ 0.010 & 0.712 $\pm$ 0.008 \\
          & Tree & 7 & 0.701 $\pm$ 0.020 & 0.705 $\pm$ 0.015 & 0.708 $\pm$ 0.017 \\
          & NN   & --& 0.707 $\pm$ 0.013 & 0.704 $\pm$ 0.020 & 0.714 $\pm$ 0.012 \\
\bottomrule
\end{tabular}
\caption{Test accuracy of classifiers using the logprob-only and full feature sets. Values are mean $\pm$ standard deviation over 5 runs. Bold indicates the best result per dataset.}
\label{tab:classifier-results-short}
\end{table*}

\subsection{Classifier Performance}
\label{sec: classifier performance}
To evaluate the joint predictive power of log-probabilities and occurrence-based features, we train decision trees and neural networks to classify hallucinations using both logprob-only and full-feature input settings.
For the raw frequency model, the most effective occurrence-based features are often the average prompt key phrase frequency and the full-generation frequency. For the $n$-gram model, longer $n$-grams such as 3-grams or 4-grams typically outperform shorter ones, while 5-grams tend to perform worse due to data sparsity, i.e., the exact 5-gram often does not appear in the training corpus. Table~\ref{tab:classifier-results-short} reports test accuracies across model types, averaged over five random seeds. Across datasets and feature types, incorporating occurrence-based features generally yields the best results, with the strongest improvements observed on NQ-Open, where questions are more challenging and the log-probability signal alone is less reliable. A similar trend is observed under ROUGE-L evaluation, with full results provided in Appendix~\ref{sec:appendix-em}.

\subsection{Effect of Stopword Filtering}
We applied stopword filtering to prompt-based features (see Section~\ref{sec:methodology}) and observed slight AUROC improvements, particularly for shorter $n$-grams. Classifier accuracy also improved modestly on some datasets, especially TriviaQA. Overall, however, the impact of stopword filtering was limited and did not materially change our conclusions. Full results are provided in Appendix~\ref{sec:appendix-stopword}.

\section{Analysis}
\subsection{Interpreting Occurrence Feature Utility}
To understand why the full-feature decision tree achieves higher test accuracy than the log-probability-only counterpart despite similar training performance, we conduct a bootstrap resampling analysis on NQ-Open using the EM criterion. We repeatedly train 200 depth-3 trees on resampled versions of the training data and measure the fraction of test samples that receive identical predictions across all runs (\textit{prediction consistency}). The log-probability-only model shows \textbf{zero consistency (0.00)}, indicating strong sensitivity to training perturbations, whereas the full-feature model achieves a \textbf{higher consistency of 0.06}. This improvement, though moderate in magnitude, reflects a clear directional trend: occurrence-based features lead to more reproducible decision boundaries and improved generalization stability.

\begin{figure*}[t]
    \centering
    \begin{subfigure}[t]{0.48\textwidth}
        \centering
        \includegraphics[width=\linewidth]{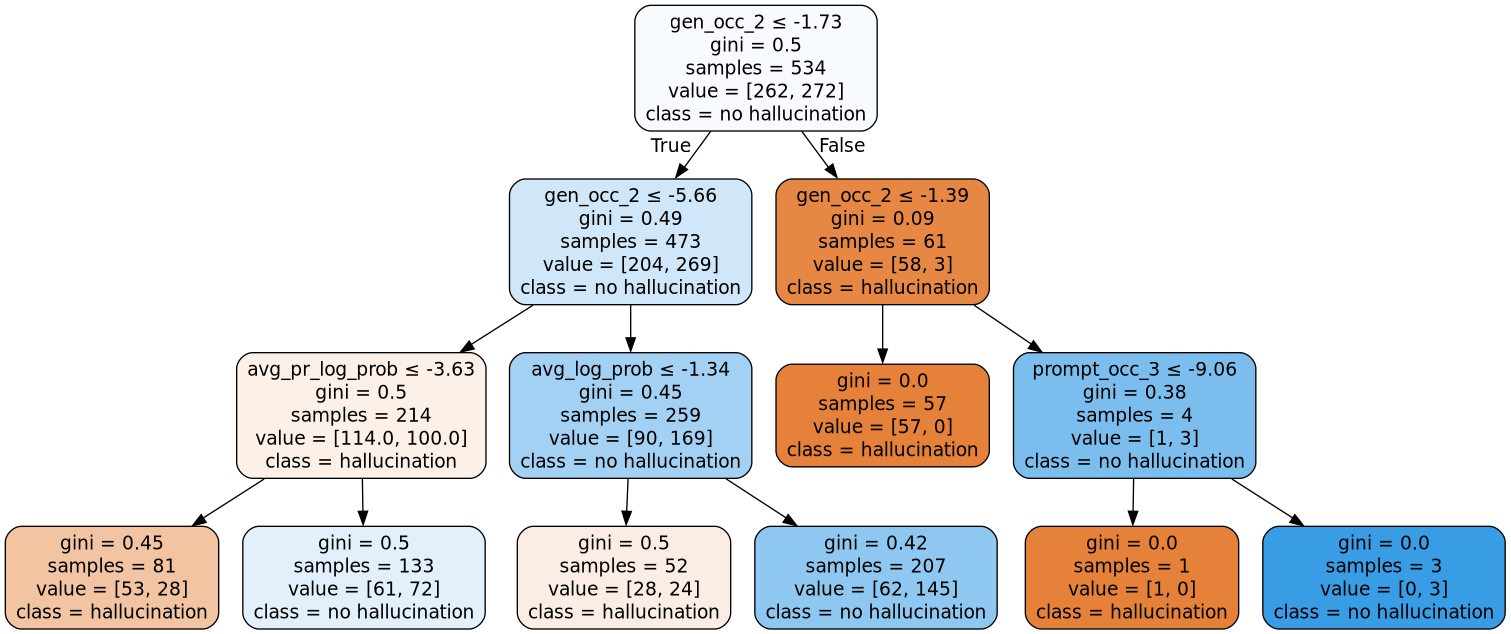}
        \caption{Full features}
        \label{fig:tree-full}
    \end{subfigure}
    \hfill
    \begin{subfigure}[t]{0.48\textwidth}
        \centering
        \includegraphics[width=\linewidth]{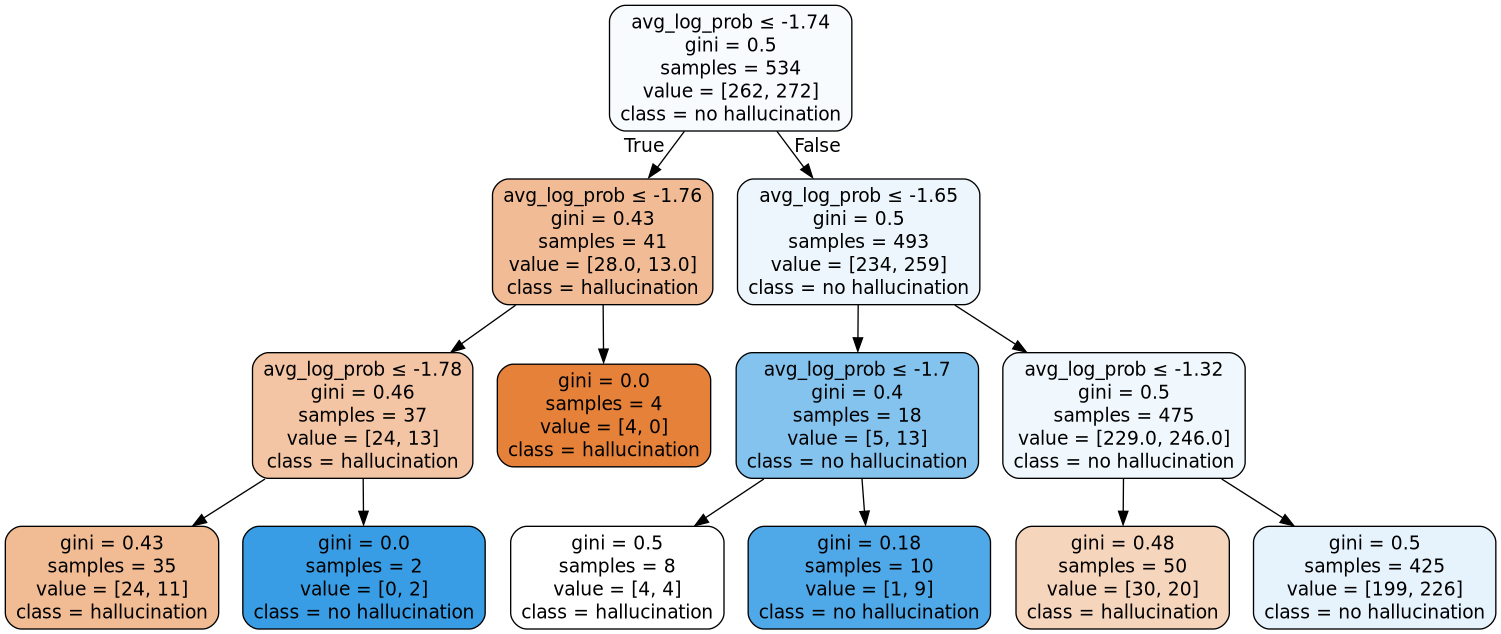}
        \caption{Log-probability only}
        \label{fig:tree-log}
    \end{subfigure}
    \caption{Decision trees (depth 3) on NQ-Open. The full-feature model (left) splits early on generation 2-gram score (\texttt{gen\_occ\_2}), isolating a hallucination-prone cluster. The log-only tree (right) lacks this separation.}
    \label{fig:tree-sidebyside}
\end{figure*}

We also identify concrete cases where occurrence features effectively differentiate hallucinated outputs.
Specifically, surface artifacts such as repeated training prompts (e.g., \texttt{Q:}) in place of an actual answer. To examine how these distinctions arise, we visualize depth-3 decision trees trained with both the full feature set and log-probabilities alone. As shown in Figure~\ref{fig:tree-sidebyside}, the tree with occurrence features cleanly separates 57 hallucinated instances along the path 
\textit{gen\_occ\_2 (generation 2-gram score)} $> -1.73$ \textbf{$\rightarrow$} \textit{gen\_occ\_2} $\leq -1.39$. 
Examples from this branch are presented in Table~\ref{tab:q_prefix_examples}.

This split captures outputs with potentially high likelihood but low lexical diversity, particularly sequences that reproduce input-style prompts rather than providing substantive answers. Such artifacts cluster within a narrow 2-gram occurrence range under the pretraining distribution, enabling occurrence-based models to flag them more consistently. These examples demonstrate that training-data-aligned features complement intrinsic uncertainty measures by revealing brittle, format-driven hallucinations that token-level probabilities alone fail to detect.

\begin{table*}[t]
\centering
\small
\resizebox{\textwidth}{!}{%
\begin{tabular}{p{0.45\linewidth} p{0.1\linewidth} p{0.22\linewidth} p{0.05\linewidth} p{0.16\linewidth}}
\toprule
\textbf{Question} & \textbf{Generation} & \textbf{Answer} & \textbf{LogP} & \textbf{2-gram score (gen)} \\
\midrule
When was the Tower of London finished being built & \texttt{Q:} & 1078 & -0.97 & -1.42 \\
How many episodes are there in Season 6 of Nashville & \texttt{Q:} & 16 & -0.38 & -1.42 \\
What is the revolution period of Venus in Earth years & \texttt{Q:} & 224.7 Earth days / 0.615 yr & -1.29 & -1.42 \\
\bottomrule
\end{tabular}
}
\caption{Examples of hallucinated generations in NQ-Open where the model outputs \texttt{Q:}. These cases fall within a specific range of generation 2-gram scores (\texttt{gen\_occ\_2}) and are captured by an early split in the decision tree.}
\label{tab:q_prefix_examples}
\end{table*}

\subsection{Are Occurrence-Based Features Pure Noise?}
\begin{figure}[H]
\centering
\includegraphics[width=0.9\columnwidth]{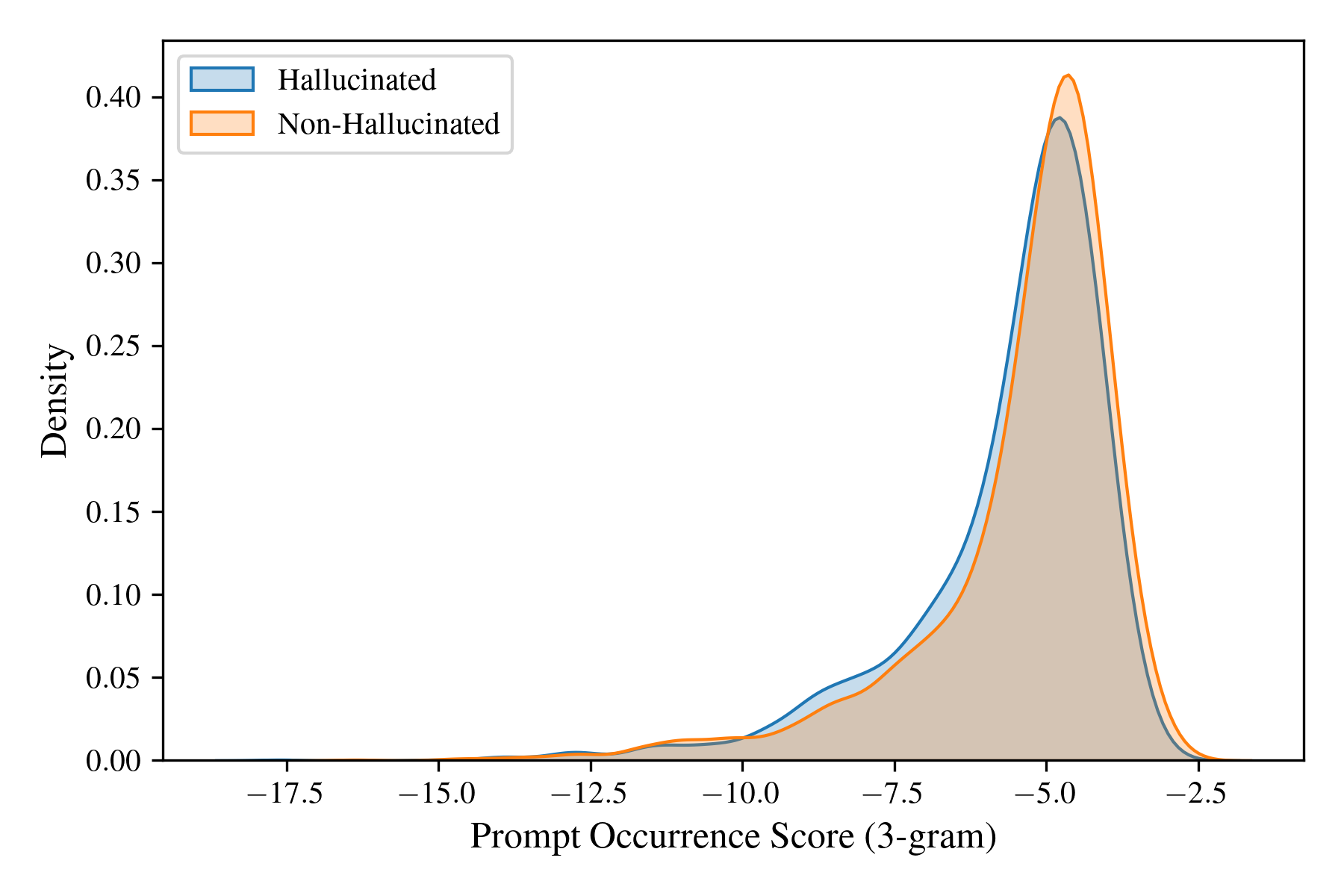}
\caption{Distribution of prompt 3-gram scores for hallucinated and non-hallucinated answers on TriviaQA.}
\label{fig:occ-dist}
\end{figure}

As shown in Figure~\ref{fig:auroc-all-em-less}, occurrence-based features generally exhibit lower AUROC scores than intrinsic model signals. To assess whether they encode any meaningful signal rather than pure noise, we visualize the distribution of prompt 3-gram occurrence scores in TriviaQA (Figure~\ref{fig:occ-dist}). The two classes--hallucinated and non-hallucinated questions--show substantial overlap yet differ modestly in their means. We ran a $t$-test confirming the difference is statistically significant ($t = -4.21$, $p = 2.7\times10^{-5}$), though given the large sample size, such tests are sensitive to even small mean shifts. This confirms that occurrence features capture a weak but consistent bias associated with hallucinated answers. While their entanglement with data frequency and format limits their standalone predictive power, they may nonetheless serve as complementary priors when combined with stronger indicators such as log-probability, as observed in Section~\ref{sec: classifier performance}.

\subsection{LLMs Are More Than Sequence Predictors: Evidence from Data Sparsity}
LLMs are often viewed as large-scale compressors of world knowledge, retrieving content memorized from training data. Our findings suggest that this view is overly reductive. Empirically, we observe that even simple natural language questions often contain 4-grams or longer spans entirely absent from a 1.3T-token pretraining corpus. For instance, $21.1\%$ of 4-grams and $44.4\%$ of 5-grams in NQ-Open do not appear anywhere in the training data. Many missing $n$-grams are not semantically obscure; for example, ``\textit{asked king to send him}''. Table~\ref{tab:data-sparsity-small} reports the percentage of missing $n$-grams across different test datasets. Detailed examples are provided in Appendix~\ref{sec: appendix-sparsity}. This indicates that models are frequently exposed to sequences they have never explicitly encountered during training, and that LLMs must generalize compositionally rather than relying solely on sequence memorization much more often than is commonly assumed.

\begin{table}[t]
\centering
\small
\resizebox{\columnwidth}{!}{
\begin{tabular}{@{}lccccc@{}}
\toprule
\textbf{Dataset} & \textbf{1-gram} & \textbf{2-gram} & \textbf{3-gram} & \textbf{4-gram} & \textbf{5-gram} \\
\midrule
TriviaQA  & 0.00 & 0.03 & 3.33 & 14.97 & 31.99 \\
NQ-Open   & 0.00 & 0.06 & 4.79 & 21.09 & 44.44 \\
CoQA      & 0.00 & 0.02 & 1.56 & 9.40  & 24.80 \\
\bottomrule
\end{tabular}
}
\caption{Percentage of $n$-grams with zero occurrences in the RedPajama training corpus across question sets from different datasets.}
\label{tab:data-sparsity-small}
\end{table}

\section{Conclusion}
We present, to our knowledge, the first systematic study of how lexical training data coverage, measured via $n$-gram frequency, relates to hallucination in LLMs. Our AUROC and classifier analyses show that while coverage-based features are weak standalone predictors, they can complement log-probabilities, especially when intrinsic confidence is unreliable. 
These results indicate that lexical-coverage features provide an additional, data-aware signal for hallucination detection tied to training exposure.

\section*{Limitations}
Due to resource constraints, our findings are based on experiments with the RedPajama-INCITE models, and may not fully generalize to larger models with more sophisticated pretraining and post-training alignments. The suffix array used to compute occurrence statistics covers 1.3 trillion tokens, which, while extensive, does not reflect the full pretraining corpora used in proprietary models. Additionally, our analysis is limited to exact $n$-gram frequency counts and does not account for paraphrasing, synonymy, or semantic similarity. Finally, our results are correlational and do not establish a causal relationship between training data exposure and hallucination behavior.

\section*{Ethics Statement}
This work studies hallucination detection in LLMs by analyzing training data coverage using $n$-gram statistics from RedPajama’s openly released 1.3T-token pretraining corpus. While ethical concerns around the large-scale use of human-authored content in LLM training are real and ongoing, particularly regarding consent, compensation, and the displacement of knowledge workers, we note that RedPajama represents a relatively transparent and responsible effort, with clear documentation and the removal of copyrighted subsets such as books. We encourage the community to continue developing tools and norms for attribution, data auditing, and fair recognition of creators whose work fuels modern LLMs.


\bibliography{custom, anthology}

\clearpage 

\appendix
\section{Appendix}
\label{sec:appendix}
\subsection{Full EM 7B Results}
\label{app: Full EM 7B Results}
\begin{figure*}[t!]
\centering

\begin{subfigure}[t]{0.49\textwidth}
    \centering
    \includegraphics[width=\linewidth]{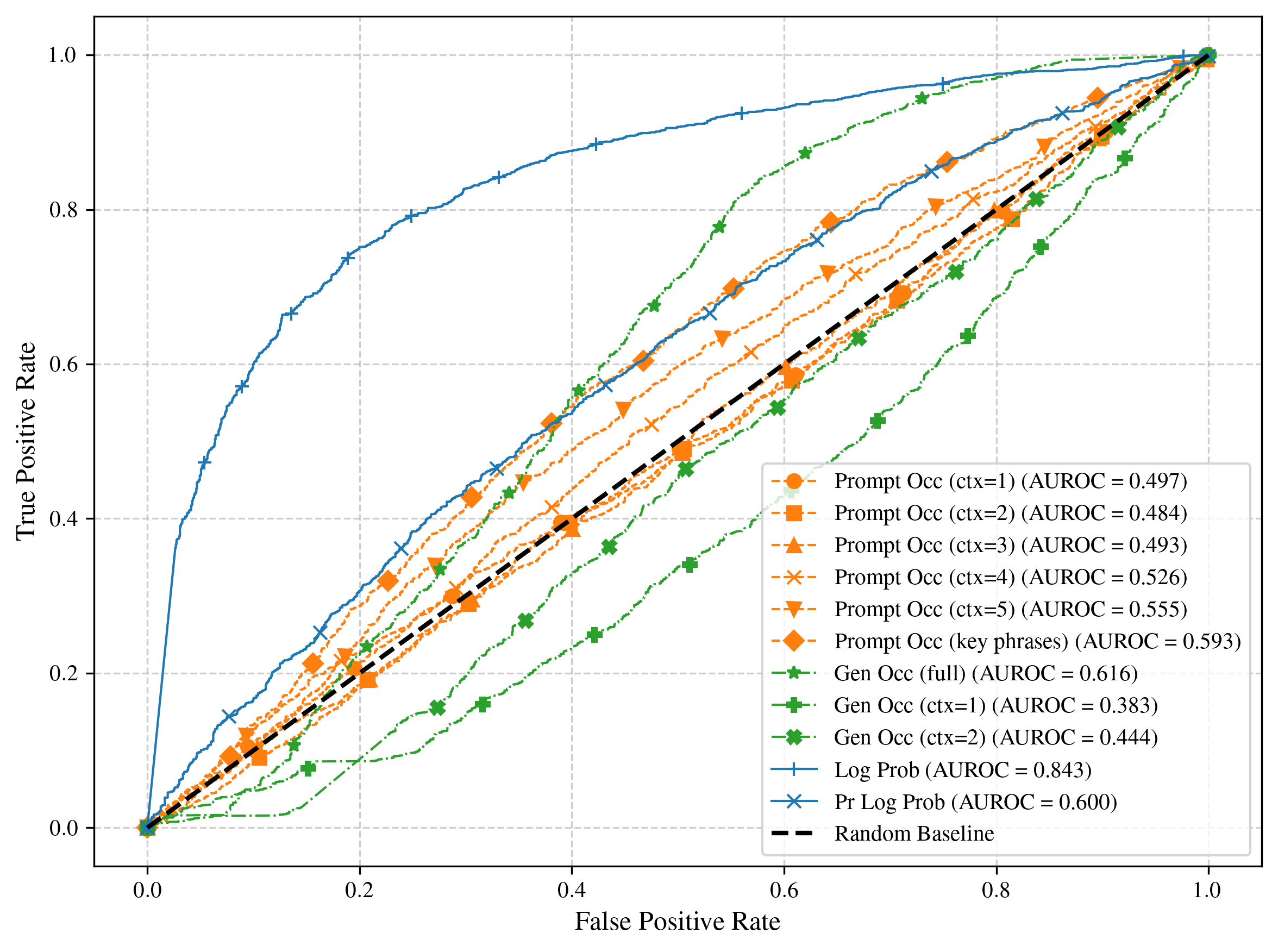}
    \caption{TriviaQA (Raw Frequency)}
\end{subfigure}
\hfill
\begin{subfigure}[t]{0.49\textwidth}
    \centering
    \includegraphics[width=\linewidth]{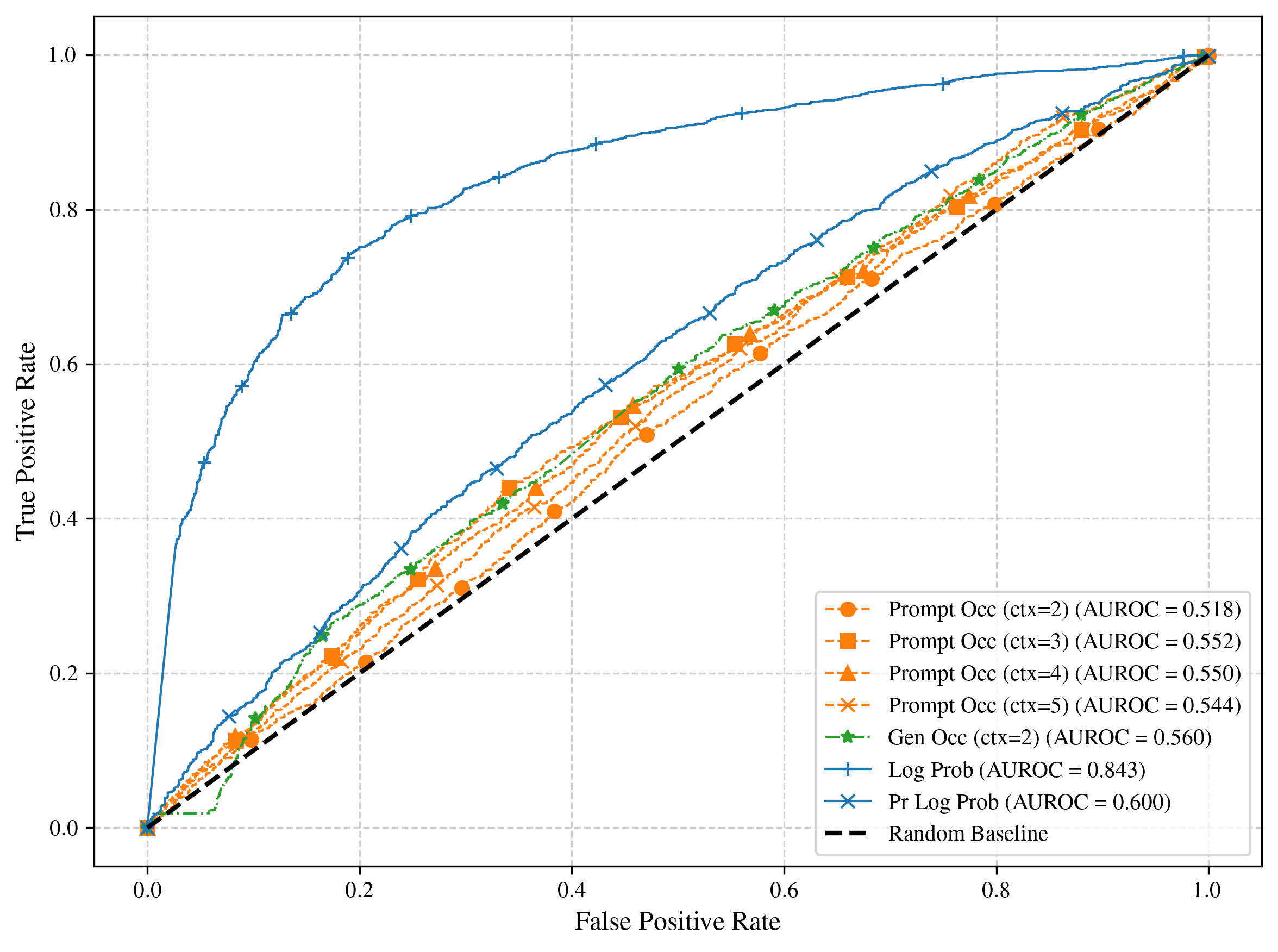}
    \caption{TriviaQA ($n$-gram Model)}
\end{subfigure}

\vspace{0.2em}

\begin{subfigure}[t]{0.49\textwidth}
    \centering
    \includegraphics[width=\linewidth]{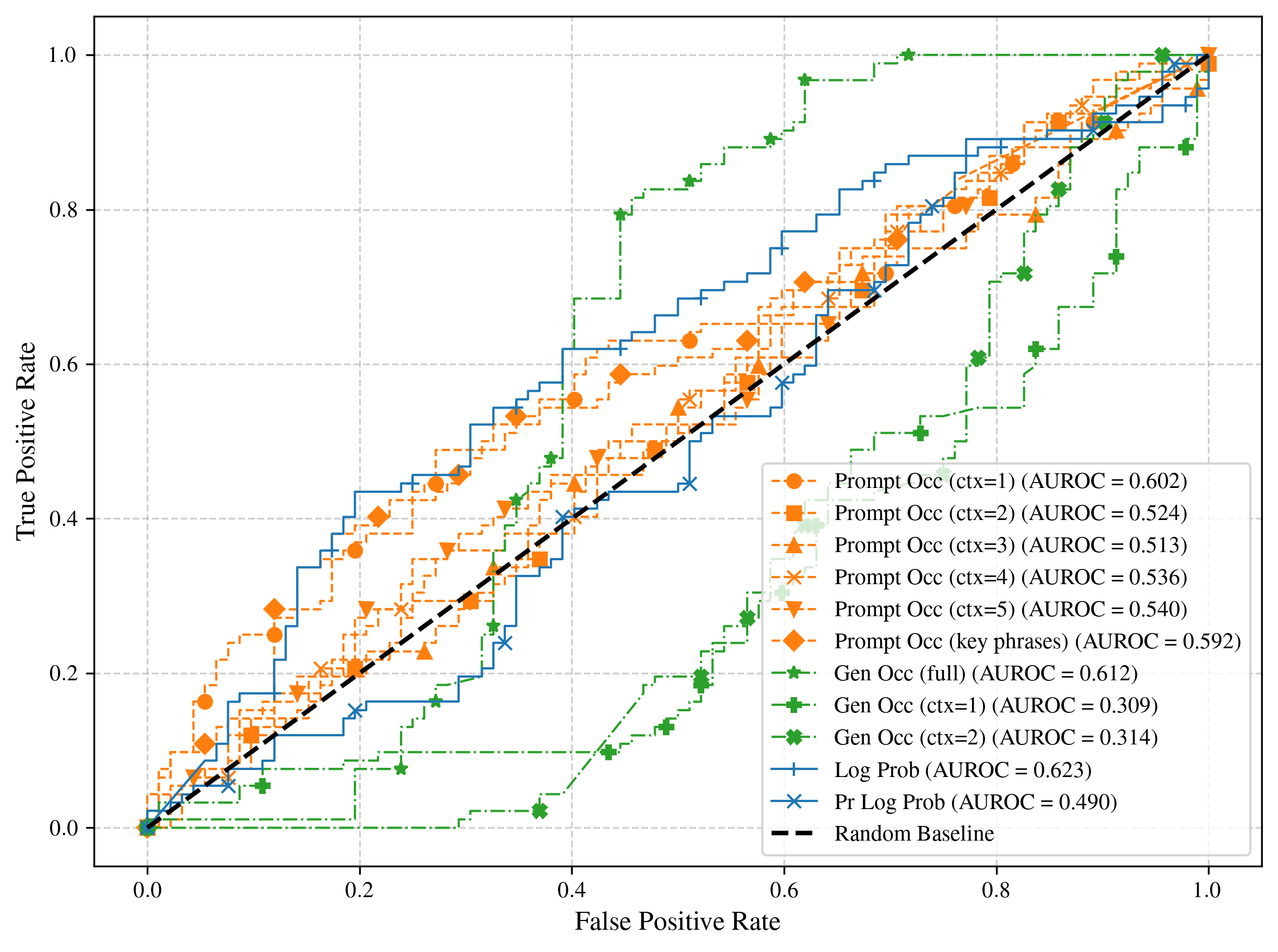}
    \caption{NQ-Open (Raw Frequency)}
\end{subfigure}
\hfill
\begin{subfigure}[t]{0.49\textwidth}
    \centering
    \includegraphics[width=\linewidth]{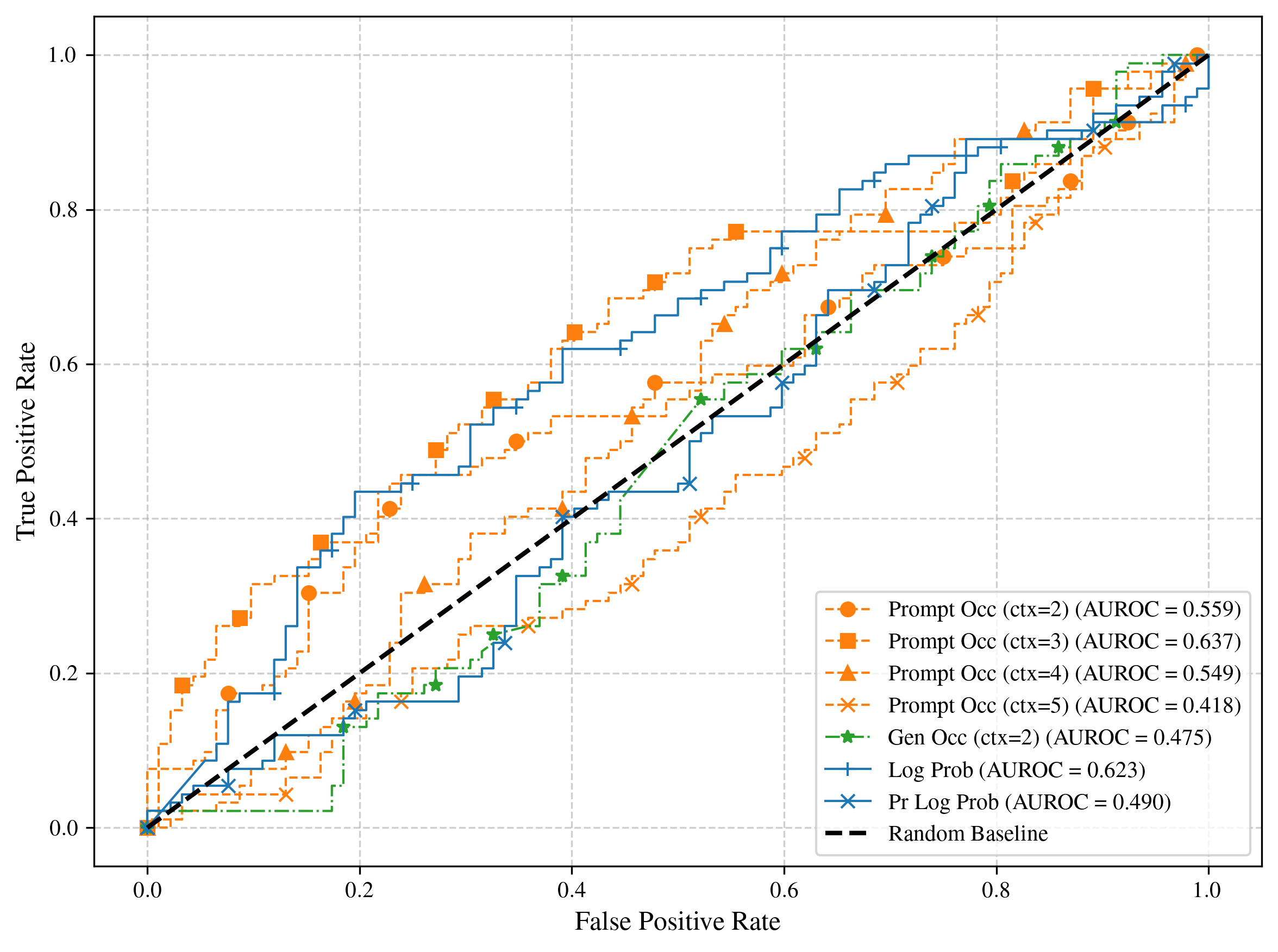}
    \caption{NQ-Open ($n$-gram Model)}
\end{subfigure}

\vspace{0.2em}

\begin{subfigure}[t]{0.49\textwidth}
    \centering
    \includegraphics[width=\linewidth]{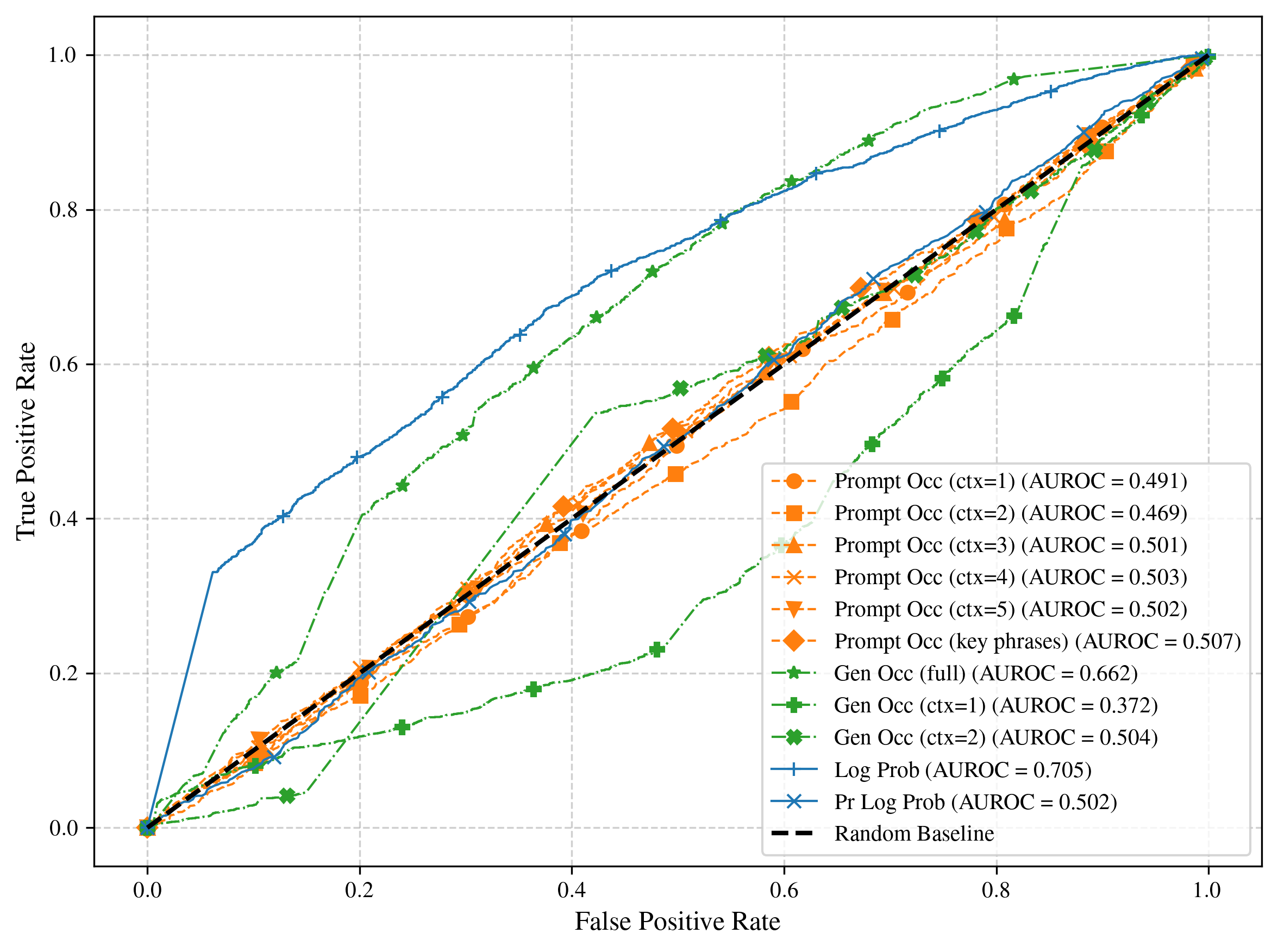}
    \caption{CoQA (Raw Frequency)}
\end{subfigure}
\hfill
\begin{subfigure}[t]{0.49\textwidth}
    \centering
    \includegraphics[width=\linewidth]{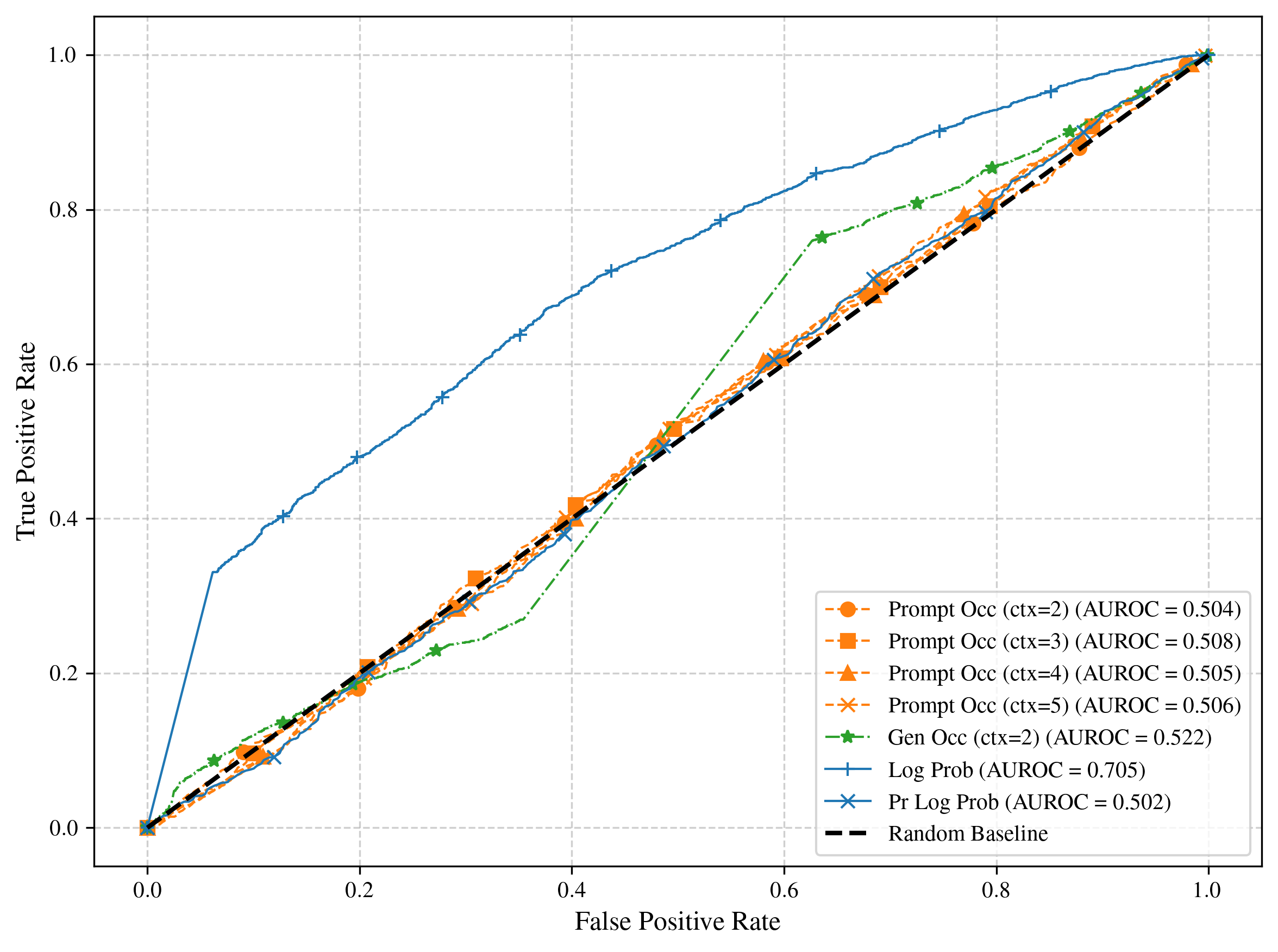}
    \caption{CoQA ($n$-gram Model)}
\end{subfigure}

\caption{AUROC curves comparing log-probabilities and occurrence-based features across datasets. Model: RedPajama-INCITE-7B; metric: EM}
\label{fig:auroc-all-em}
\end{figure*}

Figure~\ref{fig:auroc-all-em} shows the complete AUROC results, and Table~\ref{tab:acc_7b_em_new} presents the corresponding classification results for the RedPajama-7B model using EM as the evaluation metric.
\begin{table*}[t!]
\centering
\small
\resizebox{\textwidth}{!}{
\begin{tabular}{@{}lllc@{\hskip 6pt}cc|@{\hskip 6pt}cc@{}}
\toprule
\textbf{Feature Type} & \textbf{Dataset} & \textbf{Model} & \textbf{Tree Depth} & \textbf{Train Acc (Log)} & \textbf{Train Acc (Full)} & \textbf{Test Acc (Log)} & \textbf{Test Acc (Full)} \\
\midrule

\multirow{25}{*}{\textbf{Raw Freq}} & \multirow{8}{*}{TriviaQA} & \textit{Threshold} & --  & \textit{0.775 $\pm$ 0.004} & -- & \textit{0.762 $\pm$ 0.018} & -- \\
& & \multirow{6}{*}{Tree}
& 3  & 0.787 $\pm$ 0.004 & \textbf{0.796} $\pm$ 0.003 & 0.773 $\pm$ 0.026 & \textbf{0.781} $\pm$ 0.026 \\
& & & 5  & 0.799 $\pm$ 0.005 & \textbf{0.816} $\pm$ 0.002 & 0.766 $\pm$ 0.017 & \textbf{0.782} $\pm$ 0.021 \\
& & & 7  & 0.810 $\pm$ 0.005 & \textbf{0.843} $\pm$ 0.004 & 0.758 $\pm$ 0.017 & \textbf{0.778} $\pm$ 0.014 \\
& & & 9  & 0.830 $\pm$ 0.008 & \textbf{0.882} $\pm$ 0.008 & 0.757 $\pm$ 0.022 & \textbf{0.771} $\pm$ 0.015 \\
& & & 10 & 0.842 $\pm$ 0.009 & \textbf{0.902} $\pm$ 0.006 & 0.756 $\pm$ 0.029 & \textbf{0.762} $\pm$ 0.013 \\
& & & 20 & 0.945 $\pm$ 0.019 & \textbf{0.998} $\pm$ 0.002 & 0.725 $\pm$ 0.027 & \textbf{0.734} $\pm$ 0.021 \\
& & NN & -- & 0.786 $\pm$ 0.004 & \textbf{0.791} $\pm$ 0.004 & 0.776 $\pm$ 0.023 & \textbf{0.781} $\pm$ 0.018 \\

\addlinespace

& \multirow{8}{*}{NQ-Open} & \textit{Threshold} & --  & \textit{0.614 $\pm$ 0.033} & -- & \textit{0.642 $\pm$ 0.126} & -- \\
& & \multirow{6}{*}{Tree}
& 3  & 0.668 $\pm$ 0.020 & \textbf{0.836} $\pm$ 0.016 & 0.588 $\pm$ 0.076 & \textbf{0.727} $\pm$ 0.057 \\
& & & 5  & 0.726 $\pm$ 0.029 & \textbf{0.892} $\pm$ 0.019 & 0.600 $\pm$ 0.072 & \textbf{0.739} $\pm$ 0.059 \\
& & & 7  & 0.806 $\pm$ 0.032 & \textbf{0.953} $\pm$ 0.020 & 0.552 $\pm$ 0.089 & \textbf{0.733} $\pm$ 0.066 \\
& & & 9  & 0.859 $\pm$ 0.028 & \textbf{0.976} $\pm$ 0.012 & 0.552 $\pm$ 0.076 & \textbf{0.727} $\pm$ 0.061 \\
& & & 10 & 0.870 $\pm$ 0.032 & \textbf{0.983} $\pm$ 0.012 & 0.570 $\pm$ 0.087 & \textbf{0.752} $\pm$ 0.069 \\
& & & 20 & 0.965 $\pm$ 0.014 & \textbf{1.000} $\pm$ 0.000 & 0.570 $\pm$ 0.112 & \textbf{0.727} $\pm$ 0.071 \\
& & NN & -- & 0.615 $\pm$ 0.053 & \textbf{0.676} $\pm$ 0.065 & 0.546 $\pm$ 0.111 & \textbf{0.588} $\pm$ 0.102 \\

\addlinespace

& \multirow{8}{*}{CoQA} & \textit{Threshold} & --  & \textit{0.700 $\pm$ 0.004} & -- & \textit{0.697 $\pm$ 0.012} & -- \\
& & \multirow{6}{*}{Tree}
& 3  & 0.718 $\pm$ 0.006 & \textbf{0.738} $\pm$ 0.007 & 0.712 $\pm$ 0.012 & \textbf{0.728} $\pm$ 0.010 \\
& & & 5  & 0.725 $\pm$ 0.004 & \textbf{0.766} $\pm$ 0.003 & 0.709 $\pm$ 0.012 & \textbf{0.724} $\pm$ 0.021 \\
& & & 7  & 0.736 $\pm$ 0.006 & \textbf{0.797} $\pm$ 0.006 & 0.701 $\pm$ 0.020 & \textbf{0.705} $\pm$ 0.015 \\
& & & 9  & 0.751 $\pm$ 0.009 & \textbf{0.843} $\pm$ 0.010 & \textbf{0.702} $\pm$ 0.019 & 0.696 $\pm$ 0.026 \\
& & & 10 & 0.760 $\pm$ 0.010 & \textbf{0.866} $\pm$ 0.013 & \textbf{0.694} $\pm$ 0.020 & 0.688 $\pm$ 0.025 \\
& & & 20 & 0.864 $\pm$ 0.010 & \textbf{0.991} $\pm$ 0.008 & \textbf{0.677} $\pm$ 0.006 & 0.649 $\pm$ 0.024 \\
& & NN & -- & 0.711 $\pm$ 0.003 & \textbf{0.718} $\pm$ 0.004 & \textbf{0.707} $\pm$ 0.013 & 0.704 $\pm$ 0.020 \\

\midrule

\multirow{25}{*}{\textbf{N-gram}} & \multirow{8}{*}{TriviaQA} & \textit{Threshold} & --  & \textit{0.775 $\pm$ 0.004} & -- & \textit{0.762 $\pm$ 0.018} & -- \\
& & \multirow{6}{*}{Tree}
& 3  & 0.787 $\pm$ 0.004 & \textbf{0.790} $\pm$ 0.002 & \textbf{0.773} $\pm$ 0.026 & 0.772 $\pm$ 0.025 \\
& & & 5  & 0.799 $\pm$ 0.005 & \textbf{0.808} $\pm$ 0.005 & \textbf{0.766} $\pm$ 0.017 & 0.761 $\pm$ 0.034 \\
& & & 7  & 0.810 $\pm$ 0.005 & \textbf{0.835} $\pm$ 0.006 & 0.758 $\pm$ 0.017 & \textbf{0.759} $\pm$ 0.025 \\
& & & 9  & 0.830 $\pm$ 0.008 & \textbf{0.873} $\pm$ 0.009 & \textbf{0.757} $\pm$ 0.022 & 0.752 $\pm$ 0.024 \\
& & & 10 & 0.842 $\pm$ 0.009 & \textbf{0.891} $\pm$ 0.012 & \textbf{0.756} $\pm$ 0.029 & 0.747 $\pm$ 0.022 \\
& & & 20 & 0.945 $\pm$ 0.019 & \textbf{0.993} $\pm$ 0.004 & \textbf{0.725} $\pm$ 0.027 & 0.719 $\pm$ 0.017 \\
& & NN & -- & 0.786 $\pm$ 0.004 & \textbf{0.788} $\pm$ 0.006 & \textbf{0.776} $\pm$ 0.023 & 0.771 $\pm$ 0.016 \\

\addlinespace

& \multirow{8}{*}{NQ-Open} & \textit{Threshold} & --  & \textit{0.614 $\pm$ 0.033} & -- & \textit{0.642 $\pm$ 0.126} & -- \\
& & \multirow{6}{*}{Tree}
& 3  & 0.668 $\pm$ 0.020 & \textbf{0.770} $\pm$ 0.021 & \textbf{0.588} $\pm$ 0.076 & 0.552 $\pm$ 0.141 \\
& & & 5  & 0.726 $\pm$ 0.029 & \textbf{0.874} $\pm$ 0.036 & 0.600 $\pm$ 0.072 & \textbf{0.612} $\pm$ 0.135 \\
& & & 7  & 0.806 $\pm$ 0.032 & \textbf{0.939} $\pm$ 0.046 & 0.552 $\pm$ 0.089 & \textbf{0.612} $\pm$ 0.112 \\
& & & 9  & 0.859 $\pm$ 0.028 & \textbf{0.977} $\pm$ 0.027 & 0.552 $\pm$ 0.076 & \textbf{0.618} $\pm$ 0.106 \\
& & & 10 & 0.870 $\pm$ 0.032 & \textbf{0.991} $\pm$ 0.016 & 0.570 $\pm$ 0.087 & \textbf{0.594} $\pm$ 0.137 \\
& & & 20 & 0.965 $\pm$ 0.014 & \textbf{1.000} $\pm$ 0.000 & 0.570 $\pm$ 0.112 & \textbf{0.594} $\pm$ 0.143 \\
& & NN & -- & 0.615 $\pm$ 0.053 & \textbf{0.623} $\pm$ 0.048 & 0.546 $\pm$ 0.111 & \textbf{0.582} $\pm$ 0.136 \\

\addlinespace

& \multirow{8}{*}{CoQA} & \textit{Threshold} & --  & \textit{0.700 $\pm$ 0.004} & -- & \textit{0.697 $\pm$ 0.012} & -- \\
& & \multirow{6}{*}{Tree}
& 3  & 0.718 $\pm$ 0.006 & \textbf{0.724} $\pm$ 0.005 & 0.712 $\pm$ 0.012 & 0.712 $\pm$ 0.008 \\
& & & 5  & 0.725 $\pm$ 0.004 & \textbf{0.746} $\pm$ 0.008 & 0.709 $\pm$ 0.012 & \textbf{0.720} $\pm$ 0.015 \\
& & & 7  & 0.736 $\pm$ 0.006 & \textbf{0.784} $\pm$ 0.009 & 0.701 $\pm$ 0.020 & \textbf{0.708} $\pm$ 0.017 \\
& & & 9  & 0.751 $\pm$ 0.009 & \textbf{0.827} $\pm$ 0.011 & \textbf{0.702} $\pm$ 0.019 & 0.694 $\pm$ 0.005 \\
& & & 10 & 0.760 $\pm$ 0.010 & \textbf{0.849} $\pm$ 0.013 & \textbf{0.694} $\pm$ 0.020 & 0.691 $\pm$ 0.012 \\
& & & 20 & 0.864 $\pm$ 0.010 & \textbf{0.985} $\pm$ 0.009 & \textbf{0.677} $\pm$ 0.006 & 0.648 $\pm$ 0.008 \\
& & NN & -- & 0.711 $\pm$ 0.003 & \textbf{0.721} $\pm$ 0.003 & 0.707 $\pm$ 0.013 & \textbf{0.714} $\pm$ 0.012 \\

\bottomrule
\end{tabular}
}
\caption{Accuracy across tree depths, datasets, and feature types. Bold indicates higher accuracy between log-only and full features. Model: RedPajama-INCITE-7B; metric: EM.}
\label{tab:acc_7b_em_new}
\end{table*}

\subsection{Effects of Metric Used}
\label{sec:appendix-em}
\begin{figure*}[t!]
\centering

\begin{subfigure}[t]{0.49\textwidth}
    \centering
    \includegraphics[width=\linewidth]{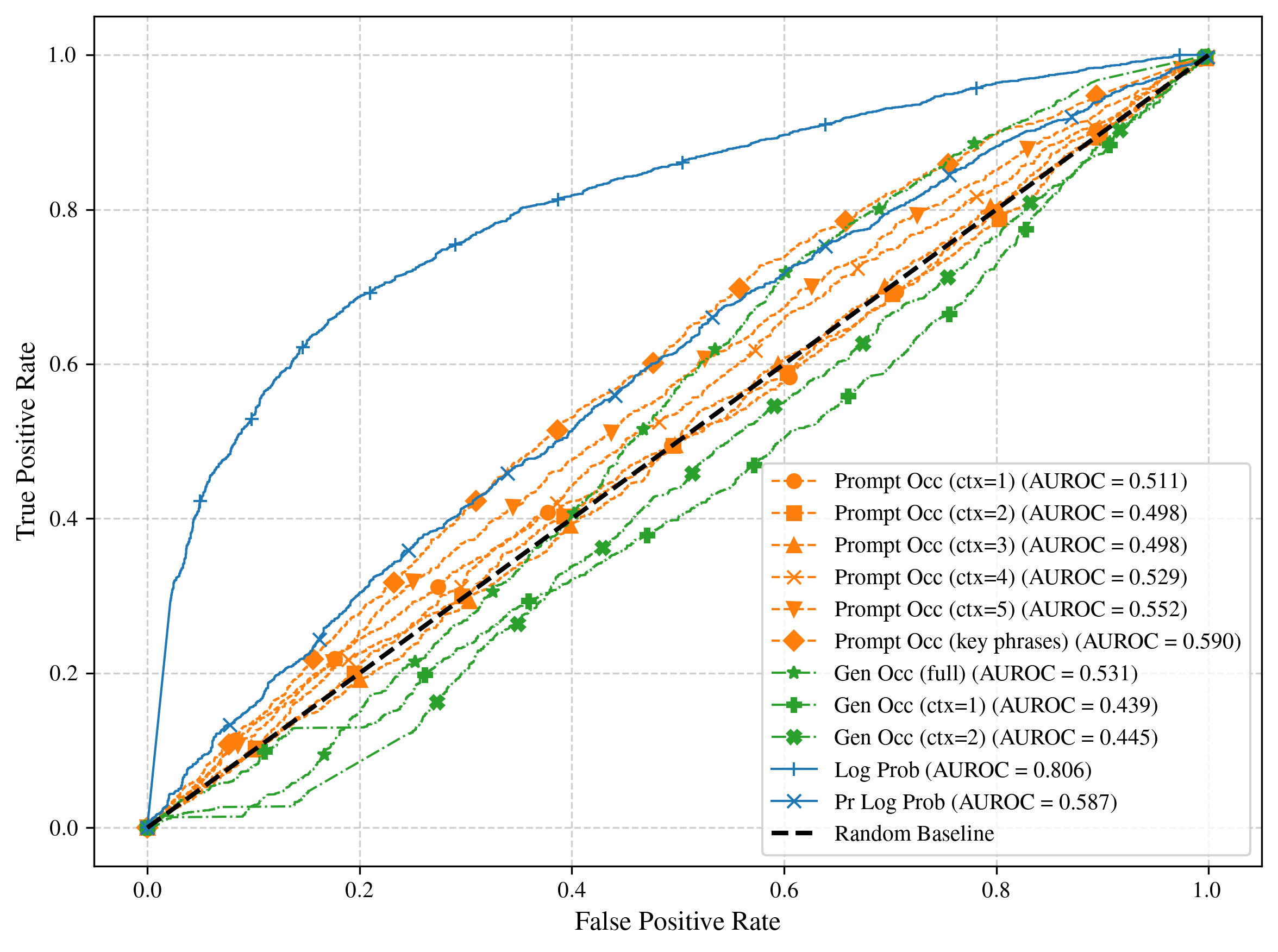}
    \caption{TriviaQA (Raw Frequency)}
\end{subfigure}
\hfill
\begin{subfigure}[t]{0.49\textwidth}
    \centering
    \includegraphics[width=\linewidth]{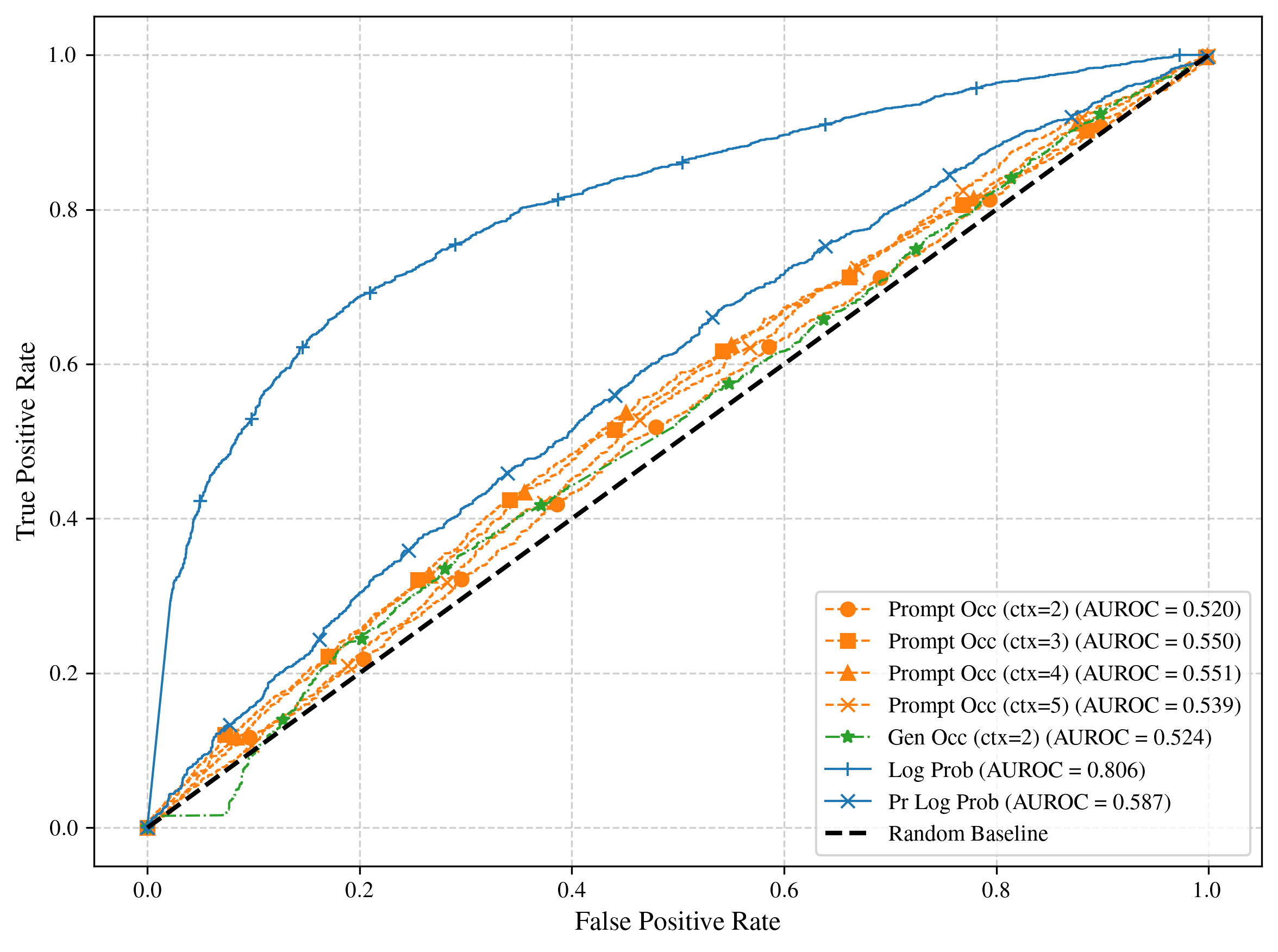}
    \caption{TriviaQA ($n$-gram Model)}
\end{subfigure}

\vspace{0.2em}

\begin{subfigure}[t]{0.49\textwidth}
    \centering
    \includegraphics[width=\linewidth]{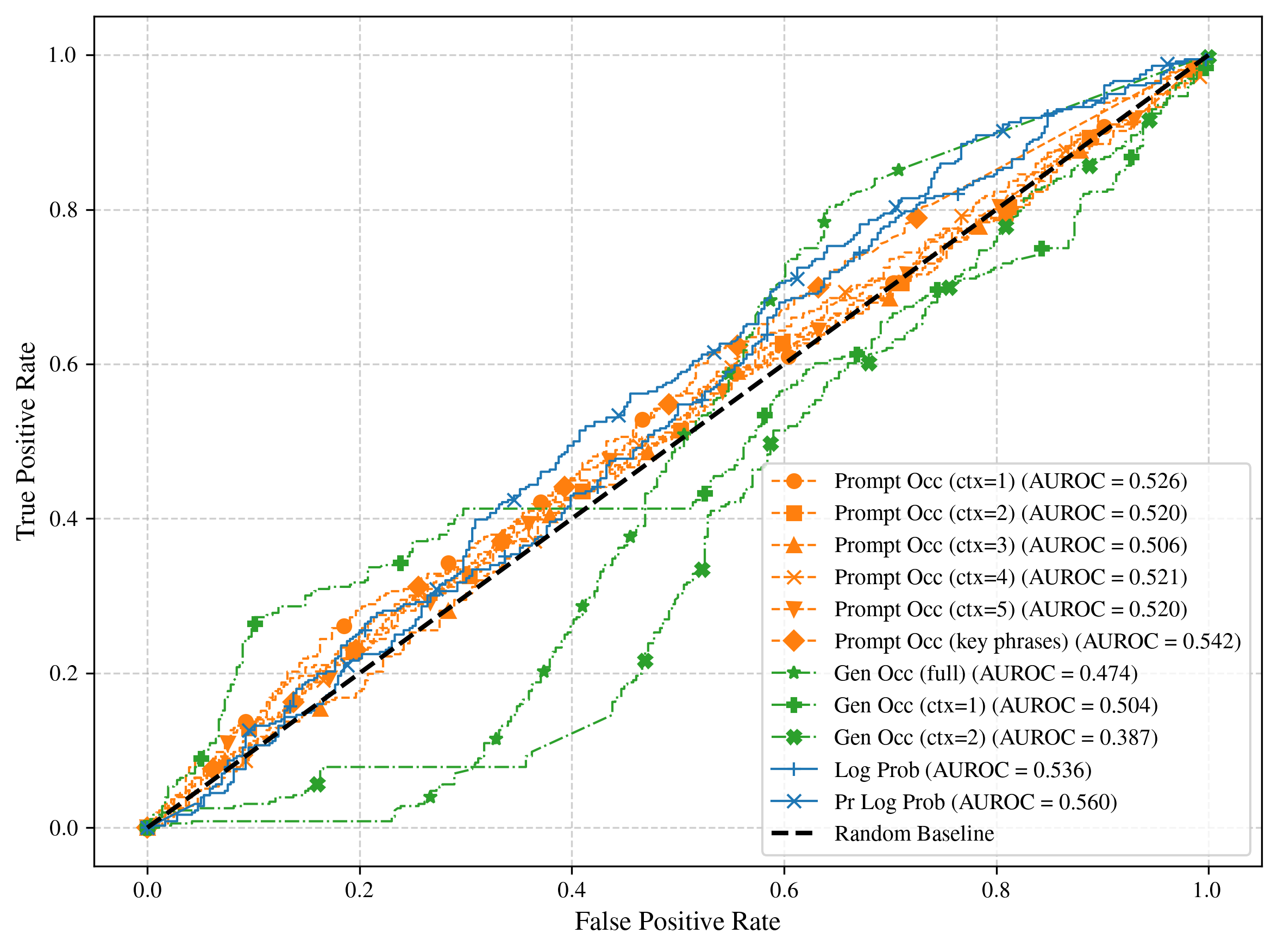}
    \caption{NQ-Open (Raw Frequency)}
\end{subfigure}
\hfill
\begin{subfigure}[t]{0.49\textwidth}
    \centering
    \includegraphics[width=\linewidth]{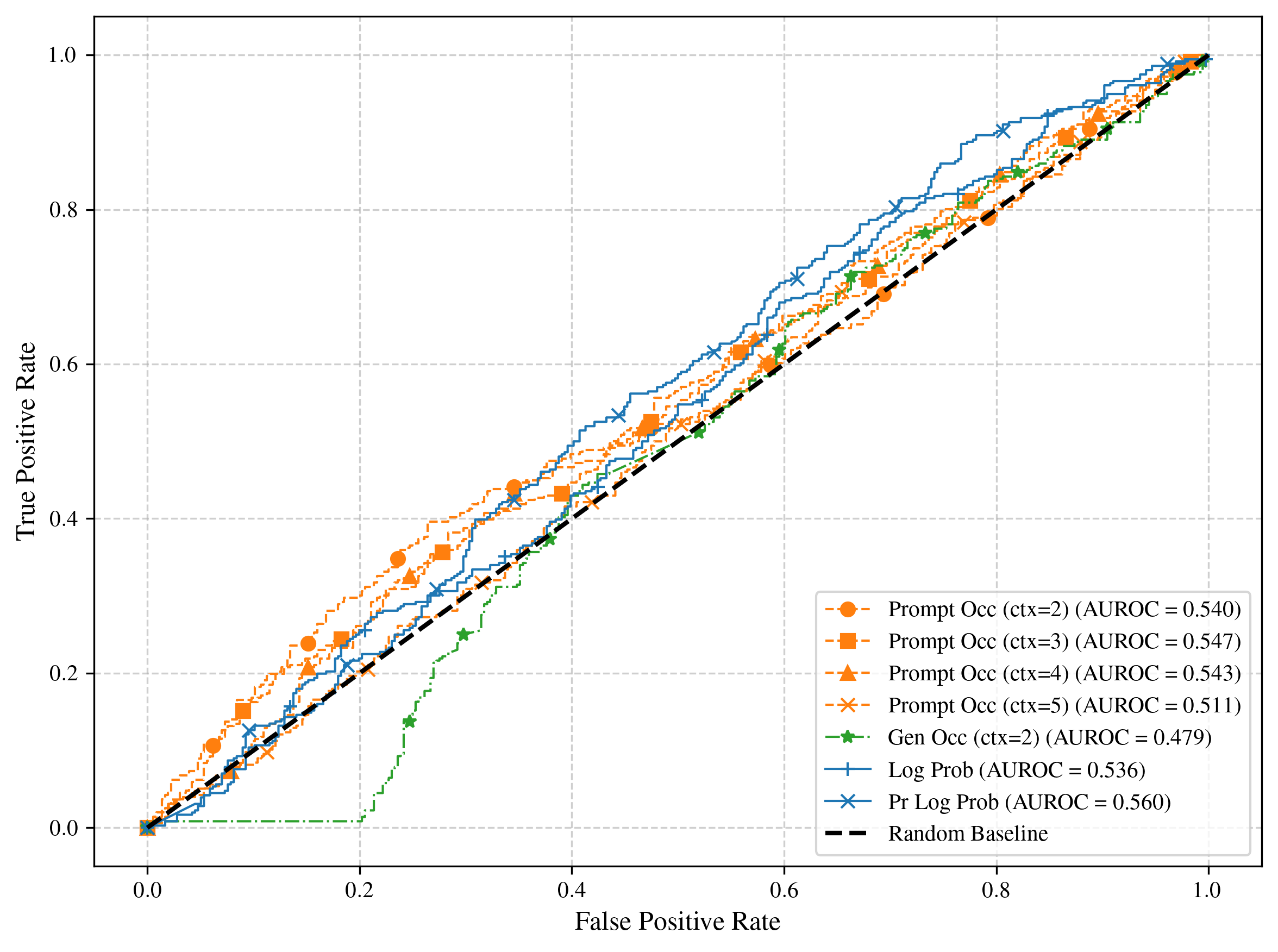}
    \caption{NQ-Open ($n$-gram Model)}
\end{subfigure}

\vspace{0.2em}

\begin{subfigure}[t]{0.49\textwidth}
    \centering
    \includegraphics[width=\linewidth]{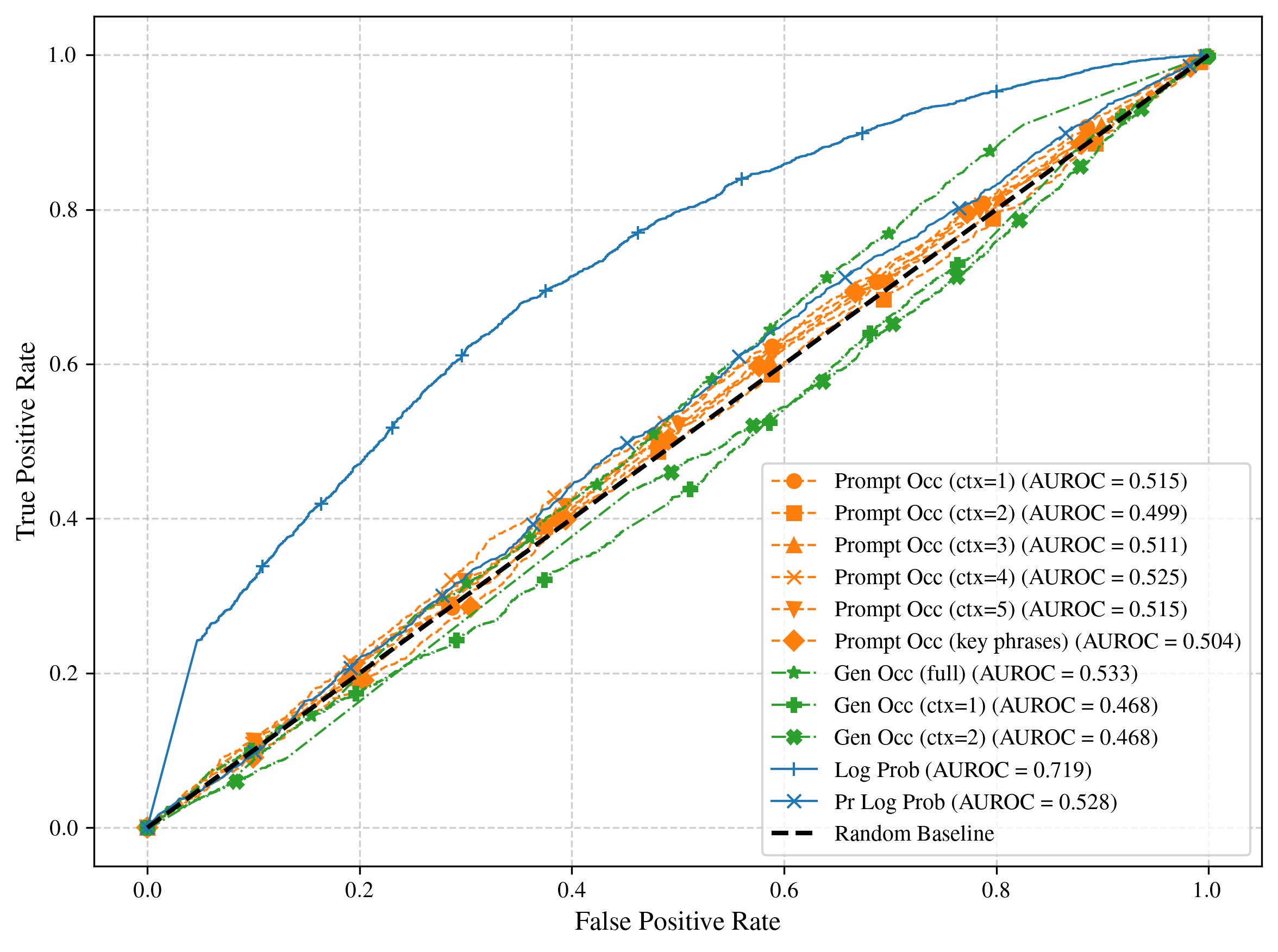}
    \caption{CoQA (Raw Frequency)}
\end{subfigure}
\hfill
\begin{subfigure}[t]{0.49\textwidth}
    \centering
    \includegraphics[width=\linewidth]{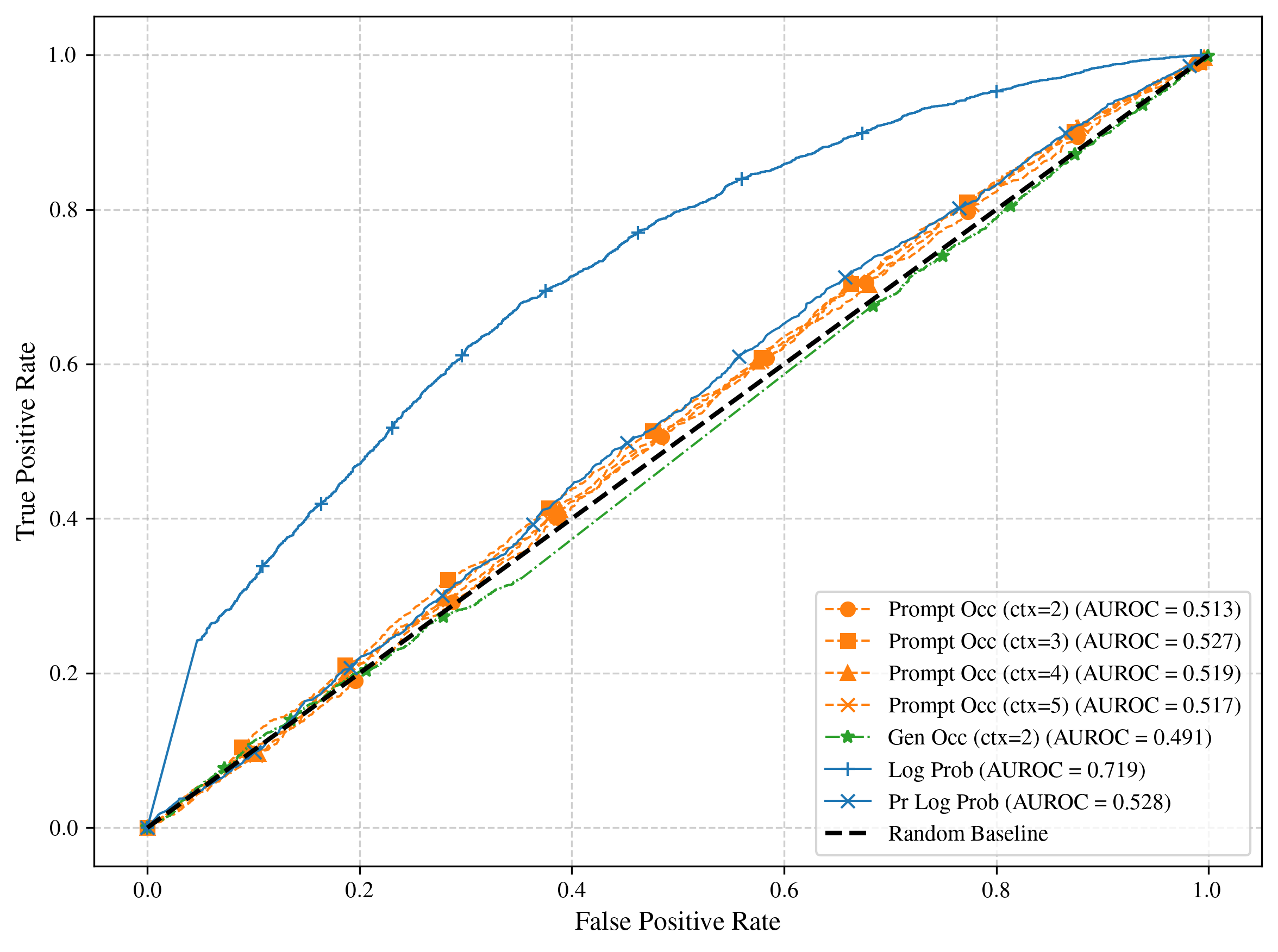}
    \caption{CoQA ($n$-gram Model)}
\end{subfigure}

\caption{AUROC curves comparing log-probabilities and occurrence-based features across datasets. Model: RedPajama-INCITE-7B; metric: ROUGE-L.}
\label{fig:auroc-all}
\end{figure*}

Figure~\ref{fig:auroc-all} reports the AUROC scores of raw frequency and $n$-gram features under ROUGE-L based labeling. The overall trends are consistent with those observed under EM, indicating that the effects of occurrence-based features are largely metric-independent. However, in extractive QA, the correlation between occurrence and faithfulness becomes noticeably weaker. Table~\ref{tab:acc_7b_rougeL_new} presents the corresponding classifier accuracy results, which follow a pattern similar to the EM based setting.
\begin{table*}[t!]
\centering
\small
\resizebox{\textwidth}{!}{
\begin{tabular}{@{}lllc@{\hskip 6pt}cc|@{\hskip 6pt}cc@{}}
\toprule
\textbf{Feature Type} & \textbf{Dataset} & \textbf{Model} & \textbf{Tree Depth} & \textbf{Train Acc (Log)} & \textbf{Train Acc (Full)} & \textbf{Test Acc (Log)} & \textbf{Test Acc (Full)} \\
\midrule

\multirow{25}{*}{\textbf{Raw Freq}} & \multirow{8}{*}{TriviaQA} & \textit{Threshold} & --  & \textit{0.728 $\pm$ 0.003} & -- & \textit{0.731 $\pm$ 0.012} & -- \\
& & \multirow{6}{*}{Tree} & 3  & 0.748 $\pm$ 0.004 & \textbf{0.752} $\pm$ 0.003 & 0.744 $\pm$ 0.015 & \textbf{0.750} $\pm$ 0.012 \\
& & & 5  & 0.753 $\pm$ 0.004 & \textbf{0.762} $\pm$ 0.004 & 0.740 $\pm$ 0.013 & \textbf{0.748} $\pm$ 0.007 \\
& & & 7  & 0.764 $\pm$ 0.002 & \textbf{0.780} $\pm$ 0.004 & 0.734 $\pm$ 0.015 & \textbf{0.741} $\pm$ 0.002 \\
& & & 9  & 0.777 $\pm$ 0.005 & \textbf{0.806} $\pm$ 0.005 & \textbf{0.733} $\pm$ 0.012 & 0.732 $\pm$ 0.004 \\
& & & 10 & 0.786 $\pm$ 0.007 & \textbf{0.824} $\pm$ 0.008 & \textbf{0.730} $\pm$ 0.010 & 0.724 $\pm$ 0.007 \\
& & & 20 & 0.892 $\pm$ 0.020 & \textbf{0.973} $\pm$ 0.016 & \textbf{0.699} $\pm$ 0.002 & 0.687 $\pm$ 0.005 \\
& & NN   & -- & 0.746 $\pm$ 0.003 & \textbf{0.752} $\pm$ 0.003 & 0.745 $\pm$ 0.013 & \textbf{0.753} $\pm$ 0.013 \\

\addlinespace

& \multirow{8}{*}{NQ-Open} & \textit{Threshold} & --  & \textit{0.536 $\pm$ 0.013} & -- & \textit{0.509 $\pm$ 0.045} & -- \\
& & \multirow{6}{*}{Tree} & 3  & 0.566 $\pm$ 0.008 & \textbf{0.682} $\pm$ 0.015 & 0.501 $\pm$ 0.045 & \textbf{0.610} $\pm$ 0.037 \\
& & & 5  & 0.603 $\pm$ 0.017 & \textbf{0.764} $\pm$ 0.007 & 0.521 $\pm$ 0.026 & \textbf{0.652} $\pm$ 0.023 \\
& & & 7  & 0.651 $\pm$ 0.014 & \textbf{0.844} $\pm$ 0.014 & 0.509 $\pm$ 0.035 & \textbf{0.636} $\pm$ 0.032 \\
& & & 9  & 0.688 $\pm$ 0.017 & \textbf{0.930} $\pm$ 0.005 & 0.487 $\pm$ 0.045 & \textbf{0.642} $\pm$ 0.015 \\
& & & 10 & 0.710 $\pm$ 0.019 & \textbf{0.956} $\pm$ 0.004 & 0.504 $\pm$ 0.044 & \textbf{0.625} $\pm$ 0.027 \\
& & & 20 & 0.883 $\pm$ 0.045 & \textbf{1.000} $\pm$ 0.000 & 0.501 $\pm$ 0.048 & \textbf{0.630} $\pm$ 0.015 \\
& & NN & -- & 0.539 $\pm$ 0.009 & \textbf{0.602} $\pm$ 0.015 & 0.518 $\pm$ 0.022 & \textbf{0.548} $\pm$ 0.056 \\

\addlinespace

& \multirow{8}{*}{CoQA} & \textit{Threshold} & --  & \textit{0.657 $\pm$ 0.003} & -- & \textit{0.654 $\pm$ 0.014} & -- \\
& & \multirow{6}{*}{Tree} & 3  & 0.662 $\pm$ 0.003 & \textbf{0.674} $\pm$ 0.002 & 0.649 $\pm$ 0.014 & \textbf{0.669} $\pm$ 0.009 \\
& & & 5  & 0.668 $\pm$ 0.003 & \textbf{0.687} $\pm$ 0.005 & 0.645 $\pm$ 0.011 & \textbf{0.657} $\pm$ 0.014 \\
& & & 7  & 0.680 $\pm$ 0.004 & \textbf{0.709} $\pm$ 0.003 & \textbf{0.643} $\pm$ 0.012 & 0.640 $\pm$ 0.007 \\
& & & 9  & 0.693 $\pm$ 0.006 & \textbf{0.746} $\pm$ 0.003 & \textbf{0.639} $\pm$ 0.009 & 0.635 $\pm$ 0.019 \\
& & & 10 & 0.701 $\pm$ 0.009 & \textbf{0.766} $\pm$ 0.007 & \textbf{0.640} $\pm$ 0.013 & 0.617 $\pm$ 0.024 \\
& & & 20 & 0.821 $\pm$ 0.023 & \textbf{0.949} $\pm$ 0.005 & \textbf{0.612} $\pm$ 0.013 & 0.573 $\pm$ 0.015 \\
& & NN & -- & 0.658 $\pm$ 0.003 & \textbf{0.660} $\pm$ 0.003 & 0.653 $\pm$ 0.016 & \textbf{0.656} $\pm$ 0.015 \\

\midrule

\multirow{25}{*}{\textbf{N-gram}} & \multirow{8}{*}{TriviaQA} & \textit{Threshold} & --  & \textit{0.728 $\pm$ 0.003} & -- & \textit{0.731 $\pm$ 0.012} & -- \\
& & \multirow{6}{*}{Tree} & 3  & 0.748 $\pm$ 0.004 & \textbf{0.751} $\pm$ 0.003 & 0.745 $\pm$ 0.015 & \textbf{0.748} $\pm$ 0.012 \\
& & & 5  & 0.754 $\pm$ 0.004 & \textbf{0.765} $\pm$ 0.005 & 0.740 $\pm$ 0.013 & \textbf{0.746} $\pm$ 0.005 \\
& & & 7  & 0.764 $\pm$ 0.003 & \textbf{0.785} $\pm$ 0.003 & 0.734 $\pm$ 0.015 & \textbf{0.742} $\pm$ 0.007 \\
& & & 9  & 0.777 $\pm$ 0.005 & \textbf{0.812} $\pm$ 0.003 & 0.733 $\pm$ 0.012 & \textbf{0.735} $\pm$ 0.007 \\
& & & 10 & 0.786 $\pm$ 0.007 & \textbf{0.826} $\pm$ 0.006 & 0.730 $\pm$ 0.010 & \textbf{0.733} $\pm$ 0.007 \\
& & & 20 & 0.892 $\pm$ 0.020 & \textbf{0.975} $\pm$ 0.013 & \textbf{0.699} $\pm$ 0.002 & 0.694 $\pm$ 0.015 \\
& & NN  & -- & 0.746 $\pm$ 0.003 & \textbf{0.748} $\pm$ 0.005 & 0.745 $\pm$ 0.013 & \textbf{0.750} $\pm$ 0.014 \\

\addlinespace

& \multirow{8}{*}{NQ-Open} & \textit{Threshold} & --  & \textit{0.536 $\pm$ 0.013} & -- & \textit{0.509 $\pm$ 0.045} & -- \\
& & \multirow{6}{*}{Tree} & 3  & 0.566 $\pm$ 0.008 & \textbf{0.661} $\pm$ 0.013 & 0.502 $\pm$ 0.045 & \textbf{0.593} $\pm$ 0.029 \\
& & & 5  & 0.603 $\pm$ 0.017 & \textbf{0.708} $\pm$ 0.016 & 0.521 $\pm$ 0.026 & \textbf{0.610} $\pm$ 0.040 \\
& & & 7  & 0.651 $\pm$ 0.014 & \textbf{0.758} $\pm$ 0.025 & 0.509 $\pm$ 0.035 & \textbf{0.613} $\pm$ 0.031 \\
& & & 9  & 0.688 $\pm$ 0.017 & \textbf{0.815} $\pm$ 0.030 & 0.487 $\pm$ 0.045 & \textbf{0.582} $\pm$ 0.038 \\
& & & 10 & 0.710 $\pm$ 0.019 & \textbf{0.843} $\pm$ 0.032 & 0.505 $\pm$ 0.044 & \textbf{0.581} $\pm$ 0.039 \\
& & & 20 & 0.883 $\pm$ 0.045 & \textbf{0.979} $\pm$ 0.016 & 0.502 $\pm$ 0.048 & \textbf{0.576} $\pm$ 0.038 \\
& & NN  & -- & 0.539 $\pm$ 0.009 & \textbf{0.611} $\pm$ 0.011 & 0.518 $\pm$ 0.022 & \textbf{0.588} $\pm$ 0.031 \\

\addlinespace

& \multirow{8}{*}{CoQA} & \textit{Threshold} & --  & \textit{0.657 $\pm$ 0.003} & -- & \textit{0.654 $\pm$ 0.014} & -- \\
& & \multirow{6}{*}{Tree} & 3  & 0.662 $\pm$ 0.003 & \textbf{0.664} $\pm$ 0.003 & \textbf{0.649} $\pm$ 0.014 & 0.647 $\pm$ 0.011 \\
& & & 5  & 0.668 $\pm$ 0.003 & \textbf{0.677} $\pm$ 0.005 & \textbf{0.645} $\pm$ 0.011 & 0.634 $\pm$ 0.010 \\
& & & 7  & 0.680 $\pm$ 0.004 & \textbf{0.700} $\pm$ 0.007 & \textbf{0.643} $\pm$ 0.012 & 0.640 $\pm$ 0.012 \\
& & & 9  & 0.693 $\pm$ 0.006 & \textbf{0.737} $\pm$ 0.016 & \textbf{0.639} $\pm$ 0.009 & 0.633 $\pm$ 0.018 \\
& & & 10 & 0.701 $\pm$ 0.009 & \textbf{0.759} $\pm$ 0.019 & \textbf{0.640} $\pm$ 0.013 & 0.627 $\pm$ 0.013 \\
& & & 20 & 0.821 $\pm$ 0.023 & \textbf{0.954} $\pm$ 0.019 & \textbf{0.612} $\pm$ 0.013 & 0.589 $\pm$ 0.014 \\
& & NN  & -- & 0.658 $\pm$ 0.003 & \textbf{0.663} $\pm$ 0.004 & 0.653 $\pm$ 0.016 & \textbf{0.660} $\pm$ 0.015 \\

\bottomrule
\end{tabular}
}
\caption{Accuracy across tree depths, datasets, and feature types. Bold indicates higher accuracy between log-only and full features. Model: RedPajama-INCITE-7B; metric: ROUGE-L.}

\label{tab:acc_7b_rougeL_new}
\end{table*}

\subsection{Effects of Stopword Filtering}
\label{sec:appendix-stopword}
\begin{figure*}[!t]
\centering

\begin{subfigure}[t]{0.49\textwidth}
    \centering
    \includegraphics[width=\linewidth]{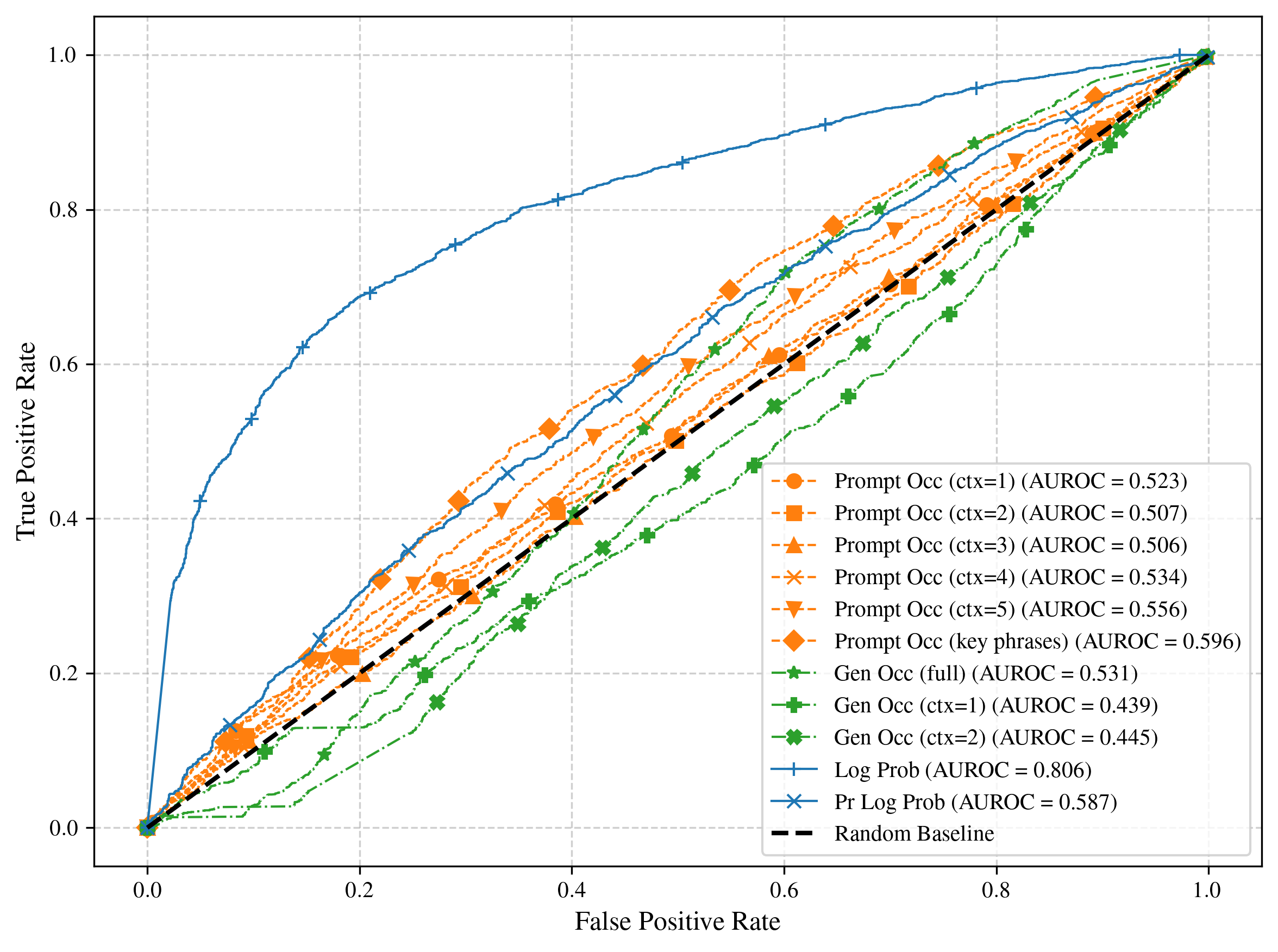}
    \caption{TriviaQA (Raw Frequency)}
\end{subfigure}
\hfill
\begin{subfigure}[t]{0.49\textwidth}
    \centering
    \includegraphics[width=\linewidth]{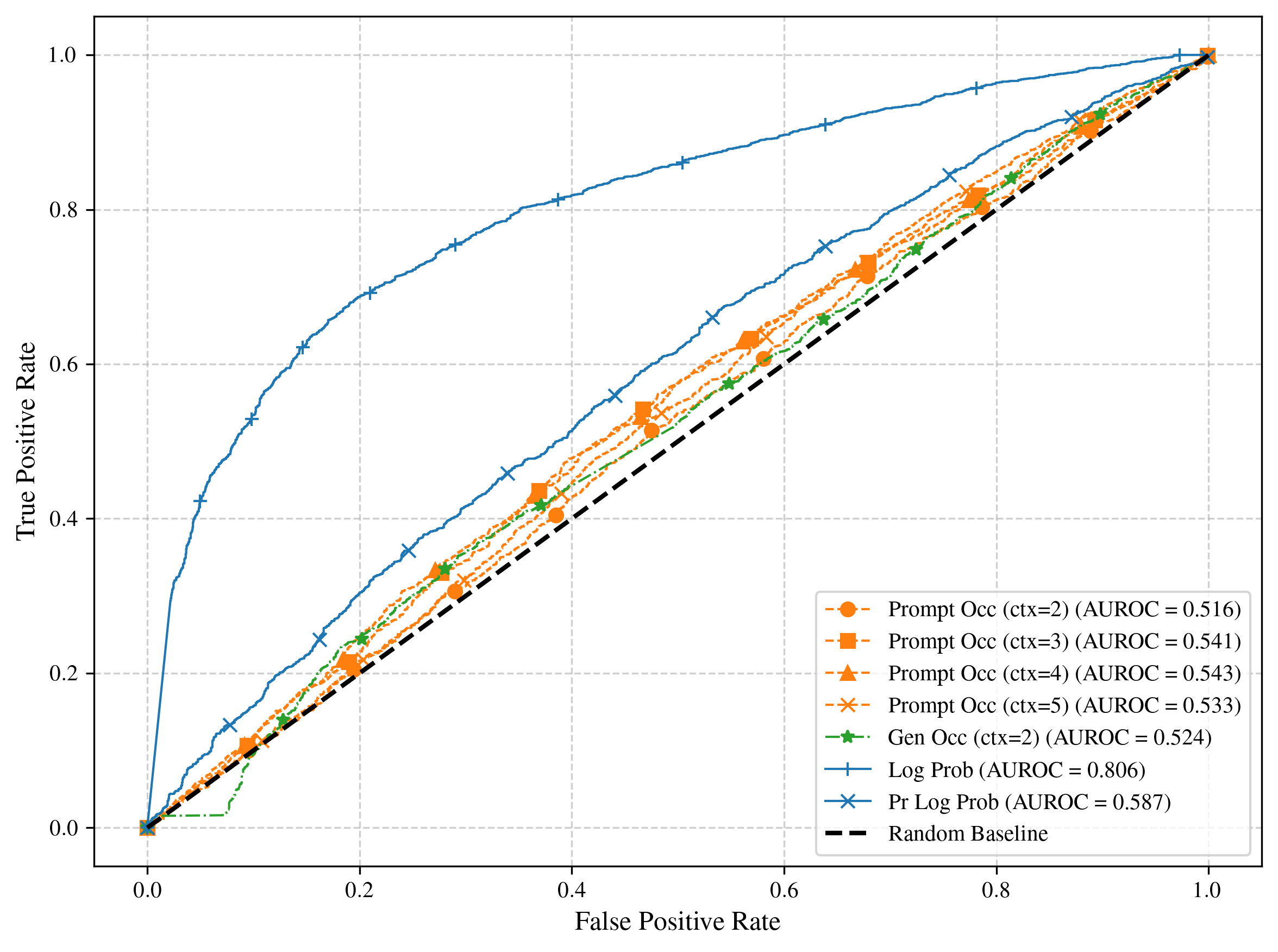}
    \caption{TriviaQA ($n$-gram Model)}
\end{subfigure}

\vspace{0.2em}

\begin{subfigure}[t]{0.49\textwidth}
    \centering
    \includegraphics[width=\linewidth]{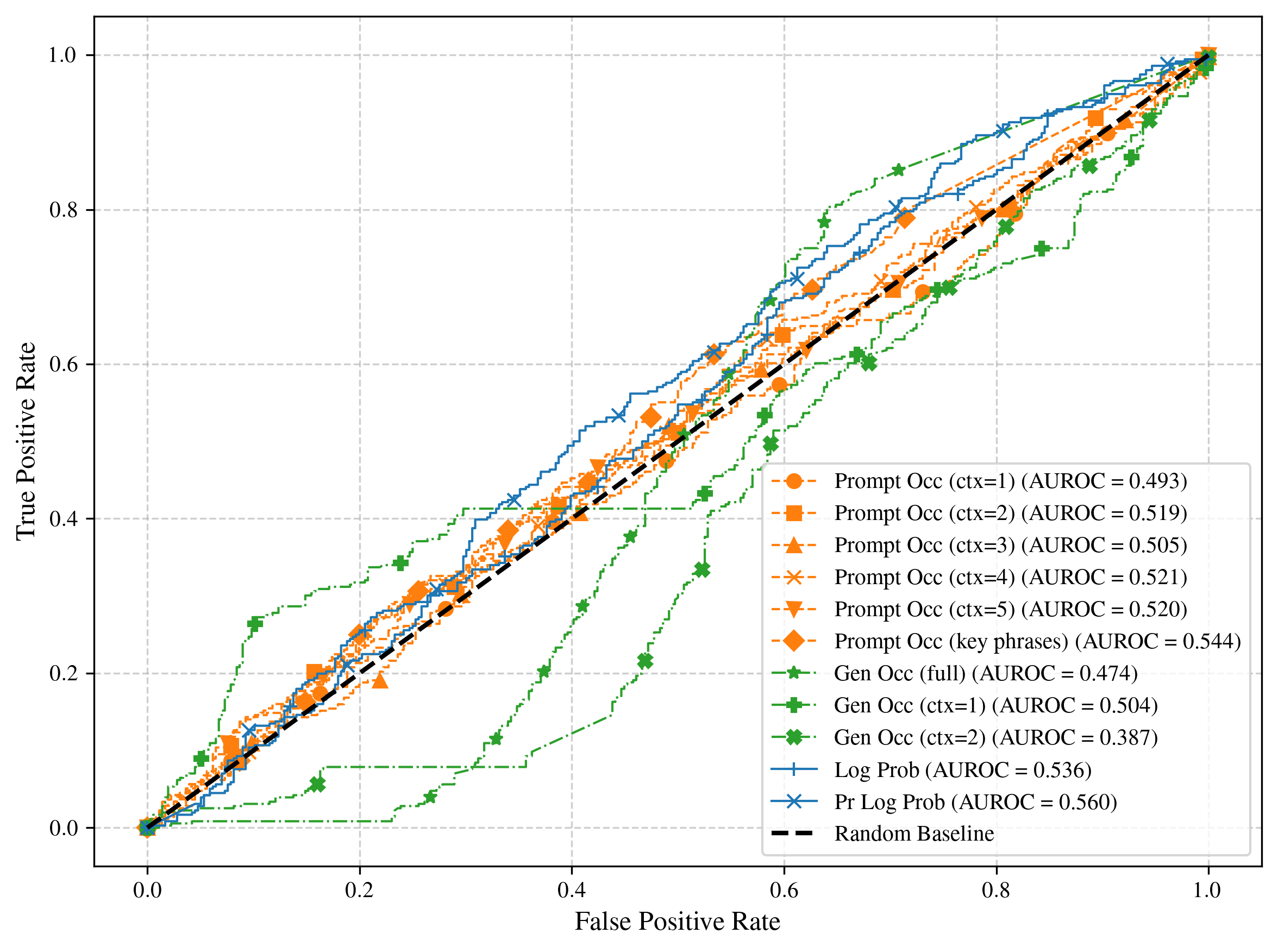}
    \caption{NQ-Open (Raw Frequency)}
\end{subfigure}
\hfill
\begin{subfigure}[t]{0.49\textwidth}
    \centering
    \includegraphics[width=\linewidth]{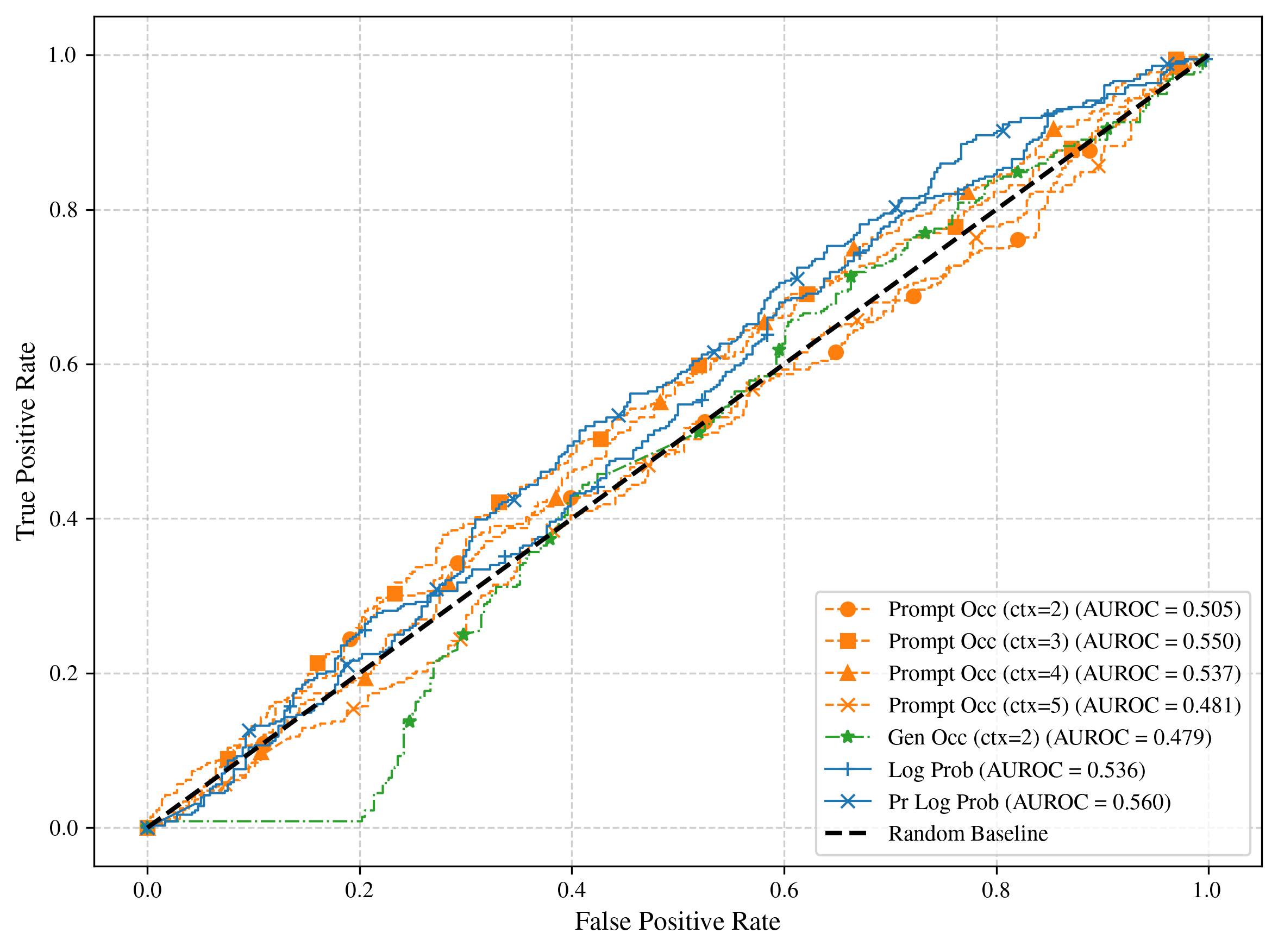}
    \caption{NQ-Open ($n$-gram Model)}
\end{subfigure}

\vspace{0.2em}

\begin{subfigure}[t]{0.49\textwidth}
    \centering
    \includegraphics[width=\linewidth]{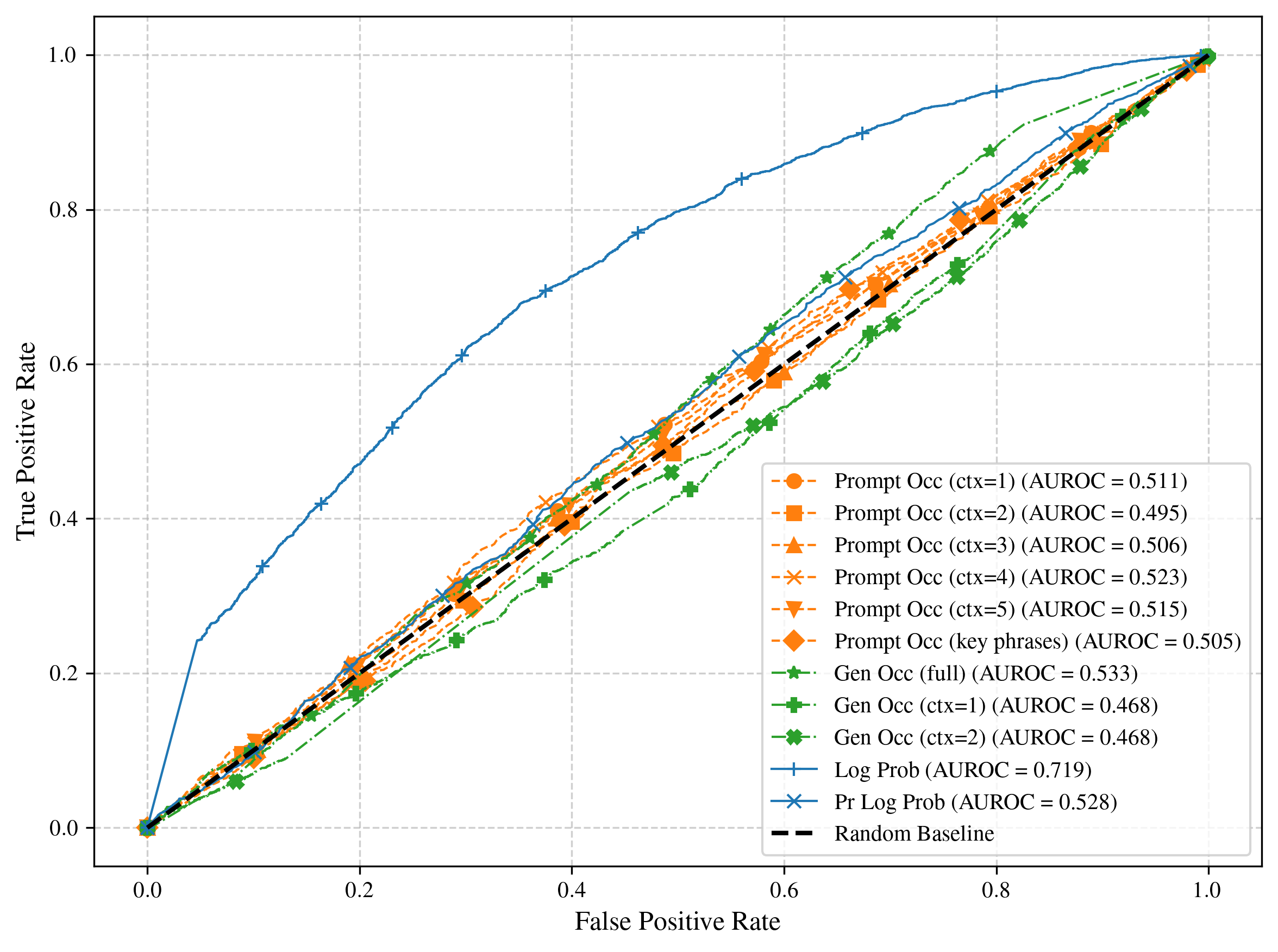}
    \caption{CoQA (Raw Frequency)}
\end{subfigure}
\hfill
\begin{subfigure}[t]{0.49\textwidth}
    \centering
    \includegraphics[width=\linewidth]{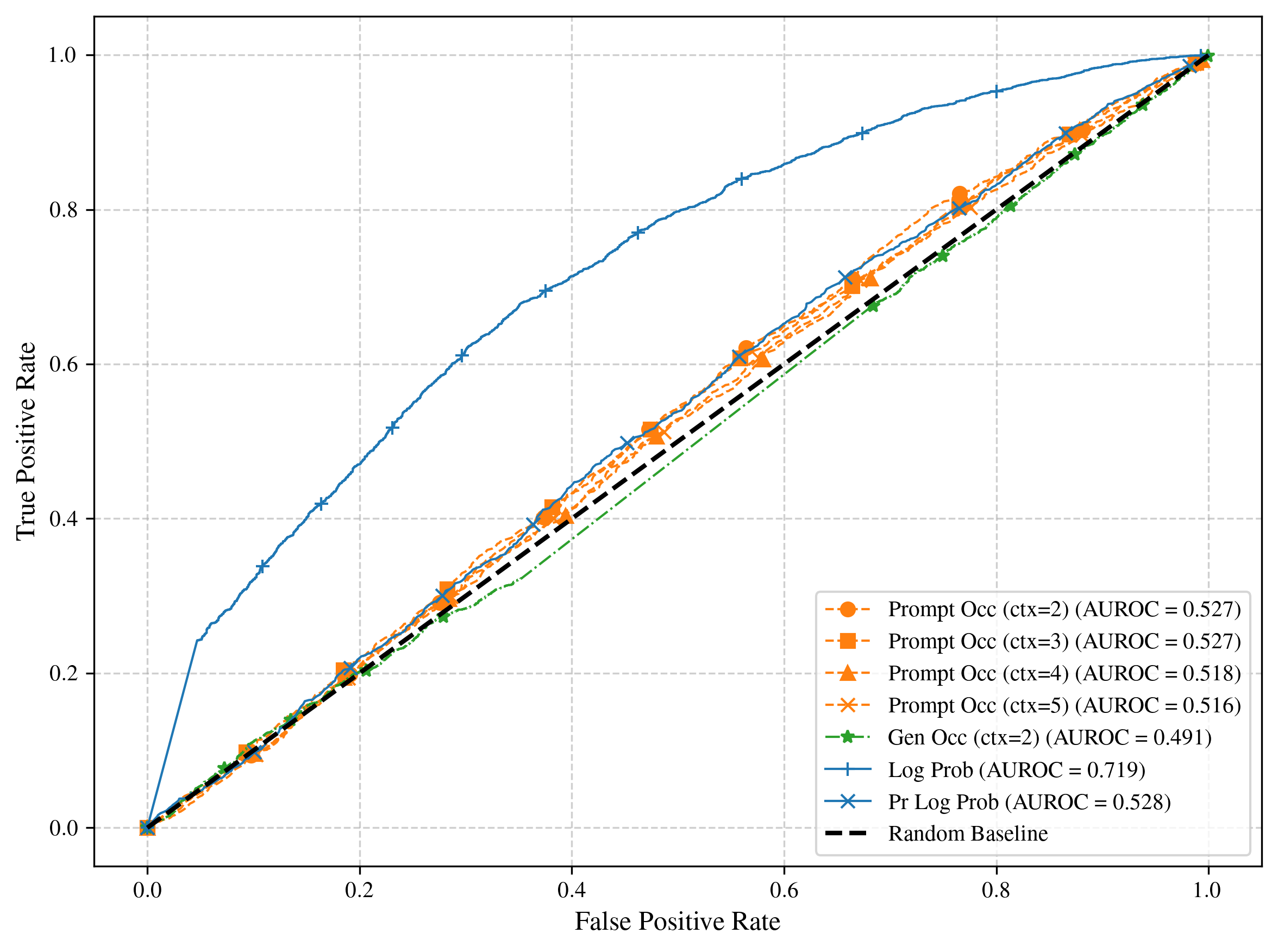}
    \caption{CoQA ($n$-gram Model)}
\end{subfigure}

\caption{AUROC curves comparing log-probabilities and occurrence-based features across datasets, with stopwords filtered. Model: RedPajama-INCITE-7B; metric: ROUGE-L}
\label{fig:auroc-all-7b-rougeL-nostopwords}
\end{figure*}

Figure~\ref{fig:auroc-all-7b-rougeL-nostopwords} shows the AUROC scores of raw frequency and $n$-gram features after applying stopword filtering on prompts, as described in Section~\ref{sec:methodology}. We observe slight improvements in AUROC, particularly for shorter $n$-grams. Table~\ref{tab:acc_7b_rougeL_nostopwords_new} presents the corresponding classifier accuracy results. Filtering leads to modest test accuracy gains, especially on TriviaQA. However, the overall impact of stopword filtering is limited and does not significantly alter our core findings.
\begin{table*}[t!]
\centering
\small
\resizebox{\textwidth}{!}{
\begin{tabular}{@{}lllc@{\hskip 6pt}cc|@{\hskip 6pt}cc@{}}
\toprule
\textbf{Feature Type} & \textbf{Dataset} & \textbf{Model} & \textbf{Tree Depth} & \textbf{Train Acc (Log)} & \textbf{Train Acc (Full)} & \textbf{Test Acc (Log)} & \textbf{Test Acc (Full)} \\
\midrule

\multirow{25}{*}{\textbf{Raw Freq}} & \multirow{8}{*}{TriviaQA} & \textit{Threshold} & --  & \textit{0.728 $\pm$ 0.003} & -- & \textit{0.731 $\pm$ 0.012} & -- \\
& & \multirow{6}{*}{Tree}
& 3  & 0.748 $\pm$ 0.004 & \textbf{0.752} $\pm$ 0.003 & 0.744 $\pm$ 0.015 & \textbf{0.750} $\pm$ 0.012 \\
& & & 5  & 0.753 $\pm$ 0.004 & \textbf{0.761} $\pm$ 0.004 & 0.740 $\pm$ 0.013 & \textbf{0.748} $\pm$ 0.007 \\
& & & 7  & 0.764 $\pm$ 0.002 & \textbf{0.782} $\pm$ 0.004 & 0.734 $\pm$ 0.015 & \textbf{0.742} $\pm$ 0.009 \\
& & & 9  & 0.777 $\pm$ 0.005 & \textbf{0.808} $\pm$ 0.007 & \textbf{0.733} $\pm$ 0.012 & 0.727 $\pm$ 0.012 \\
& & & 10 & 0.786 $\pm$ 0.007 & \textbf{0.825} $\pm$ 0.008 & \textbf{0.730} $\pm$ 0.010 & 0.725 $\pm$ 0.015 \\
& & & 20 & 0.892 $\pm$ 0.020 & \textbf{0.973} $\pm$ 0.014 & \textbf{0.699} $\pm$ 0.002 & 0.686 $\pm$ 0.019 \\
& & NN & -- & 0.746 $\pm$ 0.003 & \textbf{0.752} $\pm$ 0.003 & 0.745 $\pm$ 0.013 & \textbf{0.752} $\pm$ 0.012 \\

\addlinespace

& \multirow{8}{*}{NQ-Open} & \textit{Threshold} & --  & \textit{0.536 $\pm$ 0.013} & -- & \textit{0.509 $\pm$ 0.045} & -- \\
& & \multirow{6}{*}{Tree}
& 3  & 0.566 $\pm$ 0.008 & \textbf{0.682} $\pm$ 0.015 & 0.501 $\pm$ 0.045 & \textbf{0.610} $\pm$ 0.037 \\
& & & 5  & 0.603 $\pm$ 0.017 & \textbf{0.763} $\pm$ 0.014 & 0.521 $\pm$ 0.026 & \textbf{0.648} $\pm$ 0.019 \\
& & & 7  & 0.651 $\pm$ 0.014 & \textbf{0.841} $\pm$ 0.009 & 0.509 $\pm$ 0.035 & \textbf{0.637} $\pm$ 0.027 \\
& & & 9  & 0.688 $\pm$ 0.017 & \textbf{0.920} $\pm$ 0.010 & 0.487 $\pm$ 0.045 & \textbf{0.631} $\pm$ 0.031 \\
& & & 10 & 0.710 $\pm$ 0.019 & \textbf{0.948} $\pm$ 0.008 & 0.504 $\pm$ 0.044 & \textbf{0.619} $\pm$ 0.035 \\
& & & 20 & 0.883 $\pm$ 0.045 & \textbf{1.000} $\pm$ 0.000 & 0.501 $\pm$ 0.048 & \textbf{0.640} $\pm$ 0.045 \\
& & NN & -- & 0.539 $\pm$ 0.009 & \textbf{0.601} $\pm$ 0.014 & 0.518 $\pm$ 0.022 & \textbf{0.552} $\pm$ 0.055 \\

\addlinespace

& \multirow{8}{*}{CoQA} & \textit{Threshold} & --  & \textit{0.657 $\pm$ 0.003} & -- & \textit{0.654 $\pm$ 0.014} & -- \\
& & \multirow{6}{*}{Tree}
& 3  & 0.662 $\pm$ 0.003 & \textbf{0.674} $\pm$ 0.002 & 0.649 $\pm$ 0.014 & \textbf{0.669} $\pm$ 0.009 \\
& & & 5  & 0.668 $\pm$ 0.003 & \textbf{0.686} $\pm$ 0.007 & 0.645 $\pm$ 0.011 & \textbf{0.656} $\pm$ 0.014 \\
& & & 7  & 0.680 $\pm$ 0.004 & \textbf{0.710} $\pm$ 0.004 & 0.643 $\pm$ 0.012 & 0.643 $\pm$ 0.006 \\
& & & 9  & 0.693 $\pm$ 0.006 & \textbf{0.747} $\pm$ 0.005 & \textbf{0.639} $\pm$ 0.009 & 0.634 $\pm$ 0.011 \\
& & & 10 & 0.701 $\pm$ 0.009 & \textbf{0.768} $\pm$ 0.007 & \textbf{0.640} $\pm$ 0.013 & 0.624 $\pm$ 0.016 \\
& & & 20 & 0.821 $\pm$ 0.023 & \textbf{0.966} $\pm$ 0.011 & \textbf{0.612} $\pm$ 0.013 & 0.582 $\pm$ 0.011 \\
& & NN & -- & 0.658 $\pm$ 0.003 & \textbf{0.660} $\pm$ 0.003 & 0.653 $\pm$ 0.016 & \textbf{0.656} $\pm$ 0.015 \\

\midrule

\multirow{25}{*}{\textbf{N-gram}} & \multirow{8}{*}{TriviaQA} & \textit{Threshold} & --  & \textit{0.728 $\pm$ 0.003} & -- & \textit{0.731 $\pm$ 0.012} & -- \\
& & \multirow{6}{*}{Tree}
& 3  & 0.748 $\pm$ 0.004 & \textbf{0.751} $\pm$ 0.003 & 0.745 $\pm$ 0.015 & \textbf{0.748} $\pm$ 0.012 \\
& & & 5  & 0.754 $\pm$ 0.004 & \textbf{0.764} $\pm$ 0.004 & 0.740 $\pm$ 0.013 & \textbf{0.747} $\pm$ 0.009 \\
& & & 7  & 0.764 $\pm$ 0.003 & \textbf{0.786} $\pm$ 0.004 & 0.734 $\pm$ 0.015 & \textbf{0.744} $\pm$ 0.017 \\
& & & 9  & 0.777 $\pm$ 0.005 & \textbf{0.811} $\pm$ 0.008 & 0.733 $\pm$ 0.012 & \textbf{0.734} $\pm$ 0.015 \\
& & & 10 & 0.786 $\pm$ 0.007 & \textbf{0.828} $\pm$ 0.007 & \textbf{0.730} $\pm$ 0.010 & 0.723 $\pm$ 0.011 \\
& & & 20 & 0.892 $\pm$ 0.020 & \textbf{0.970} $\pm$ 0.010 & \textbf{0.699} $\pm$ 0.002 & 0.685 $\pm$ 0.005 \\
& & NN & -- & 0.746 $\pm$ 0.003 & \textbf{0.749} $\pm$ 0.004 & 0.745 $\pm$ 0.013 & \textbf{0.746} $\pm$ 0.014 \\

\addlinespace

& \multirow{8}{*}{NQ-Open} & \textit{Threshold} & --  & \textit{0.536 $\pm$ 0.013} & -- & \textit{0.509 $\pm$ 0.045} & -- \\
& & \multirow{6}{*}{Tree}
& 3  & 0.566 $\pm$ 0.008 & \textbf{0.665} $\pm$ 0.012 & 0.502 $\pm$ 0.045 & \textbf{0.590} $\pm$ 0.016 \\
& & & 5  & 0.603 $\pm$ 0.017 & \textbf{0.704} $\pm$ 0.017 & 0.521 $\pm$ 0.026 & \textbf{0.605} $\pm$ 0.030 \\
& & & 7  & 0.651 $\pm$ 0.014 & \textbf{0.775} $\pm$ 0.018 & 0.509 $\pm$ 0.035 & \textbf{0.606} $\pm$ 0.023 \\
& & & 9  & 0.688 $\pm$ 0.017 & \textbf{0.841} $\pm$ 0.019 & 0.487 $\pm$ 0.045 & \textbf{0.578} $\pm$ 0.023 \\
& & & 10 & 0.710 $\pm$ 0.019 & \textbf{0.869} $\pm$ 0.023 & 0.505 $\pm$ 0.044 & \textbf{0.587} $\pm$ 0.022 \\
& & & 20 & 0.883 $\pm$ 0.045 & \textbf{0.990} $\pm$ 0.019 & 0.502 $\pm$ 0.048 & \textbf{0.570} $\pm$ 0.033 \\
& & NN & -- & 0.539 $\pm$ 0.009 & \textbf{0.612} $\pm$ 0.014 & 0.518 $\pm$ 0.022 & \textbf{0.581} $\pm$ 0.028 \\

\addlinespace

& \multirow{8}{*}{CoQA} & \textit{Threshold} & --  & \textit{0.657 $\pm$ 0.003} & -- & \textit{0.654 $\pm$ 0.014} & -- \\
& & \multirow{6}{*}{Tree}
& 3  & 0.662 $\pm$ 0.003 & \textbf{0.664} $\pm$ 0.003 & \textbf{0.649} $\pm$ 0.014 & 0.647 $\pm$ 0.011 \\
& & & 5  & 0.668 $\pm$ 0.003 & \textbf{0.678} $\pm$ 0.006 & \textbf{0.645} $\pm$ 0.011 & 0.635 $\pm$ 0.012 \\
& & & 7  & 0.680 $\pm$ 0.004 & \textbf{0.702} $\pm$ 0.007 & \textbf{0.643} $\pm$ 0.012 & 0.642 $\pm$ 0.016 \\
& & & 9  & 0.693 $\pm$ 0.006 & \textbf{0.740} $\pm$ 0.007 & \textbf{0.639} $\pm$ 0.009 & 0.628 $\pm$ 0.020 \\
& & & 10 & 0.701 $\pm$ 0.009 & \textbf{0.763} $\pm$ 0.013 & \textbf{0.640} $\pm$ 0.013 & 0.625 $\pm$ 0.020 \\
& & & 20 & 0.821 $\pm$ 0.023 & \textbf{0.958} $\pm$ 0.025 & \textbf{0.612} $\pm$ 0.013 & 0.581 $\pm$ 0.017 \\
& & NN & -- & 0.658 $\pm$ 0.003 & \textbf{0.663} $\pm$ 0.005 & 0.653 $\pm$ 0.016 & \textbf{0.661} $\pm$ 0.017 \\

\bottomrule
\end{tabular}
}
\caption{Accuracy across tree depths, datasets, and feature types, with stopwords filtered from occurrence features. Model: RedPajama-INCITE-7B; metric: ROUGE-L.}

\label{tab:acc_7b_rougeL_nostopwords_new}
\end{table*}

\subsection{Effects of Model Size}
\begin{figure*}[t!]
\centering

\begin{subfigure}[t]{0.49\textwidth}
    \centering
    \includegraphics[width=\linewidth]{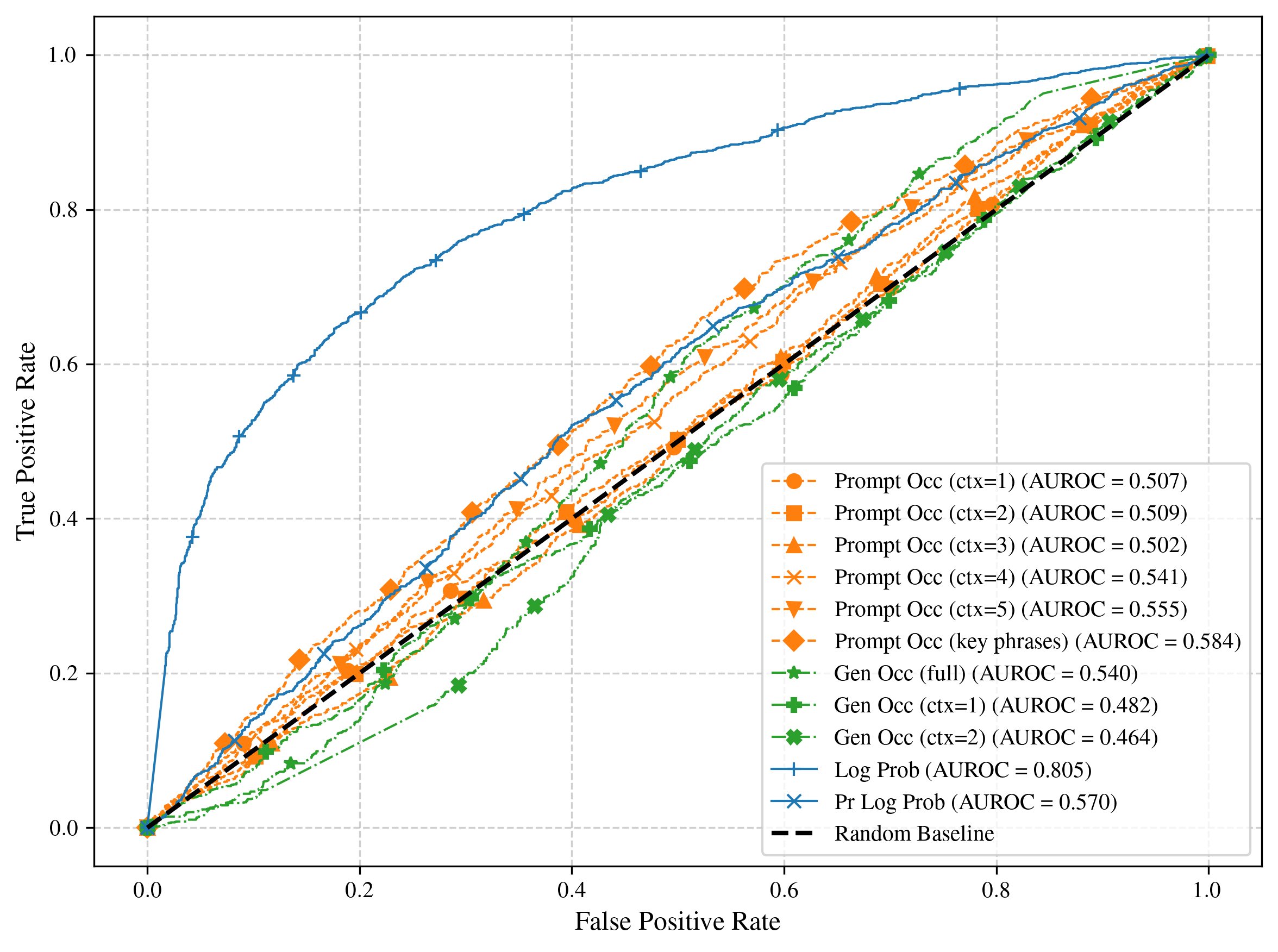}
    \caption{TriviaQA (Raw Frequency)}
\end{subfigure}
\hfill
\begin{subfigure}[t]{0.49\textwidth}
    \centering
    \includegraphics[width=\linewidth]{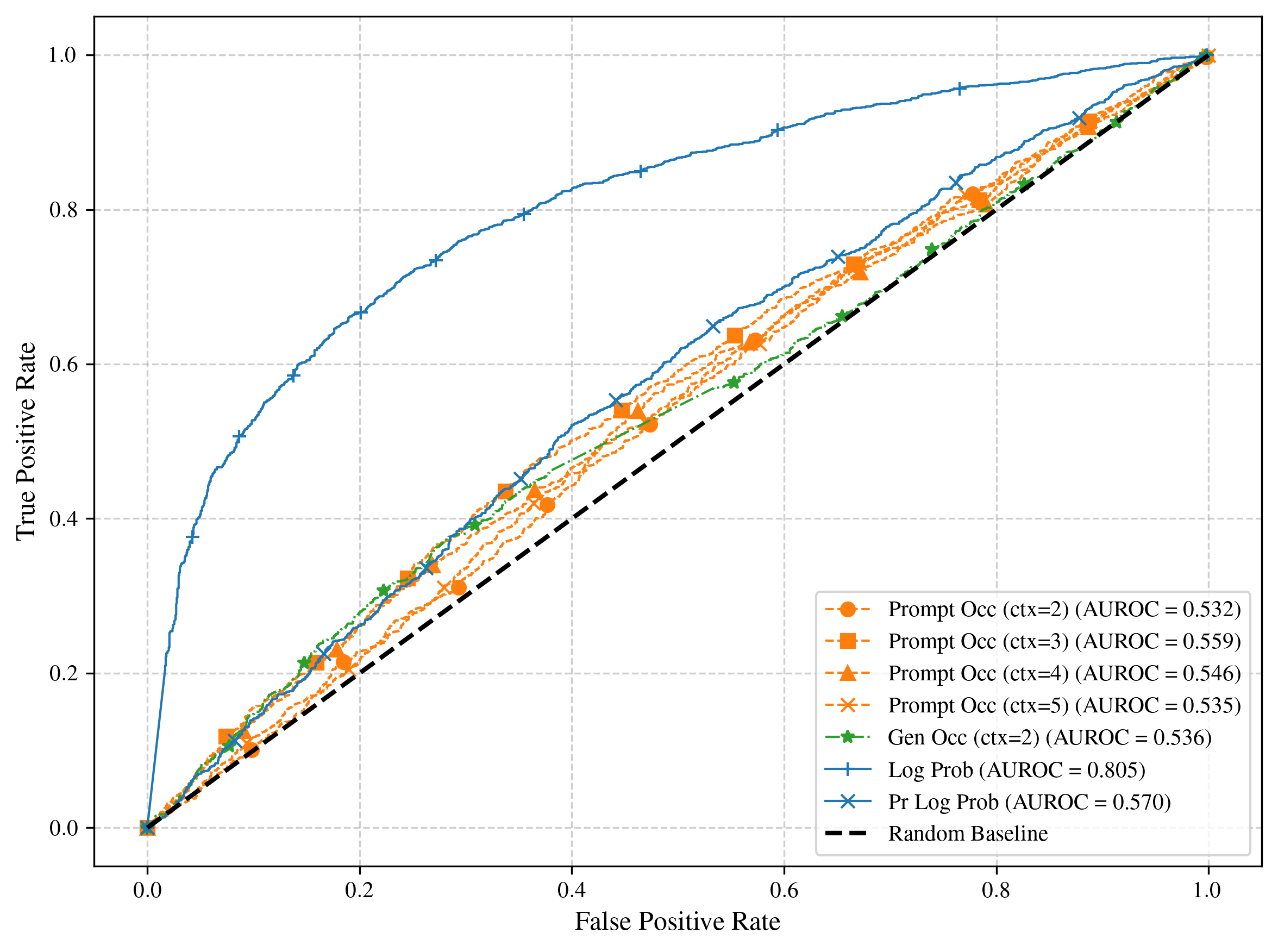}
    \caption{TriviaQA ($n$-gram Model)}
\end{subfigure}

\vspace{0.2em}

\begin{subfigure}[t]{0.49\textwidth}
    \centering
    \includegraphics[width=\linewidth]{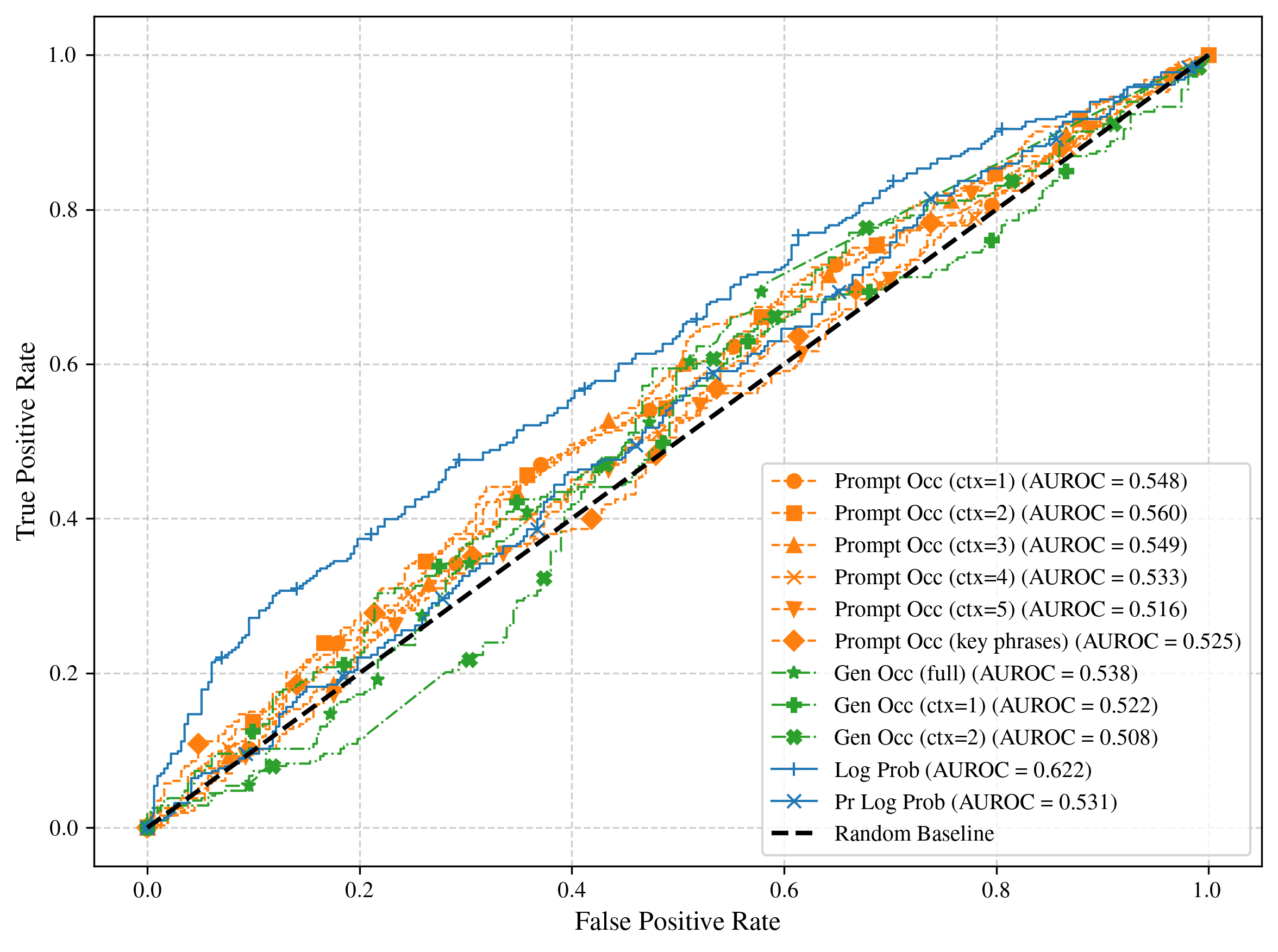}
    \caption{NQ-Open (Raw Frequency)}
\end{subfigure}
\hfill
\begin{subfigure}[t]{0.49\textwidth}
    \centering
    \includegraphics[width=\linewidth]{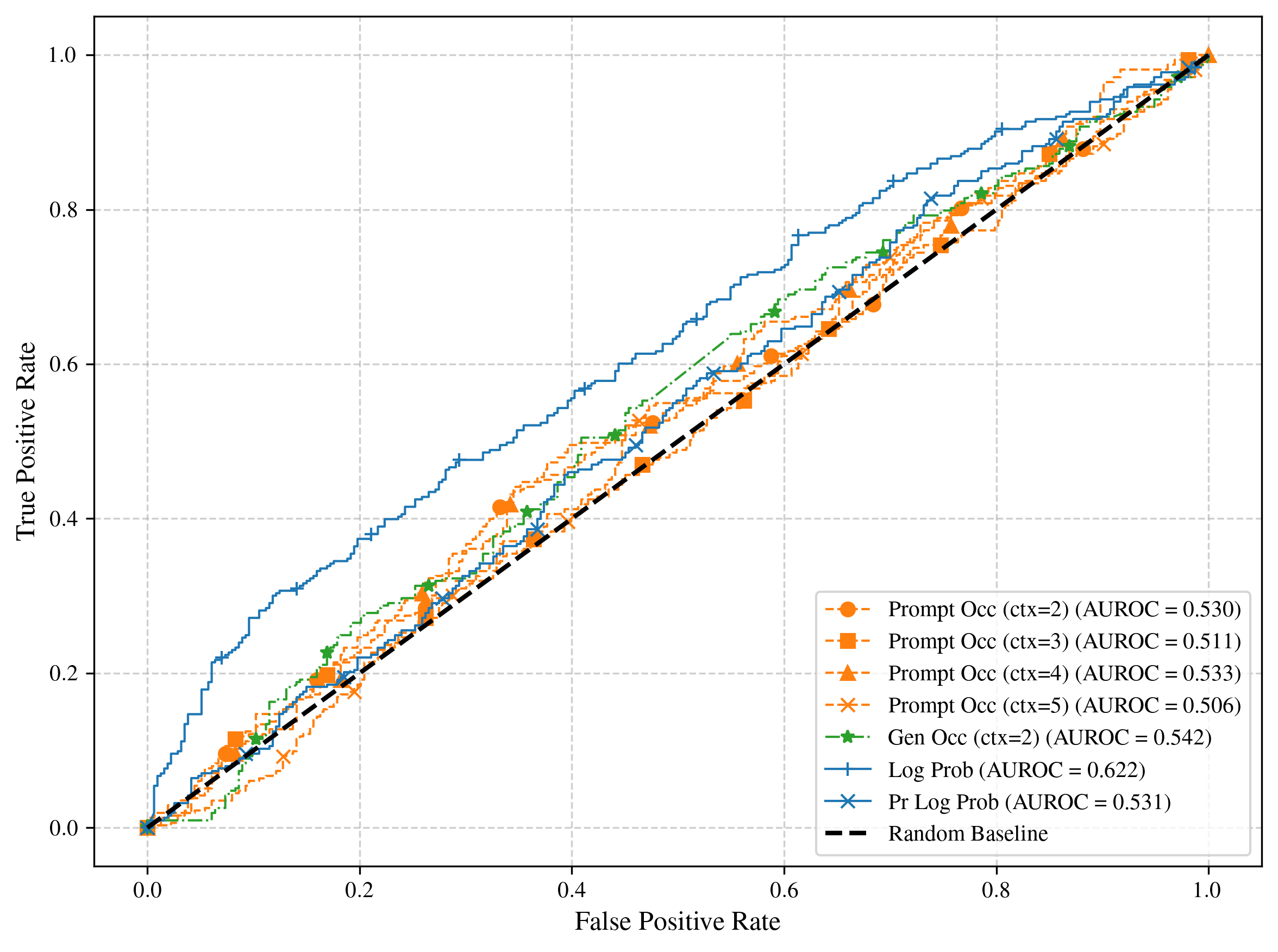}
    \caption{NQ-Open ($n$-gram Model)}
\end{subfigure}

\vspace{0.2em}

\begin{subfigure}[t]{0.49\textwidth}
    \centering
    \includegraphics[width=\linewidth]{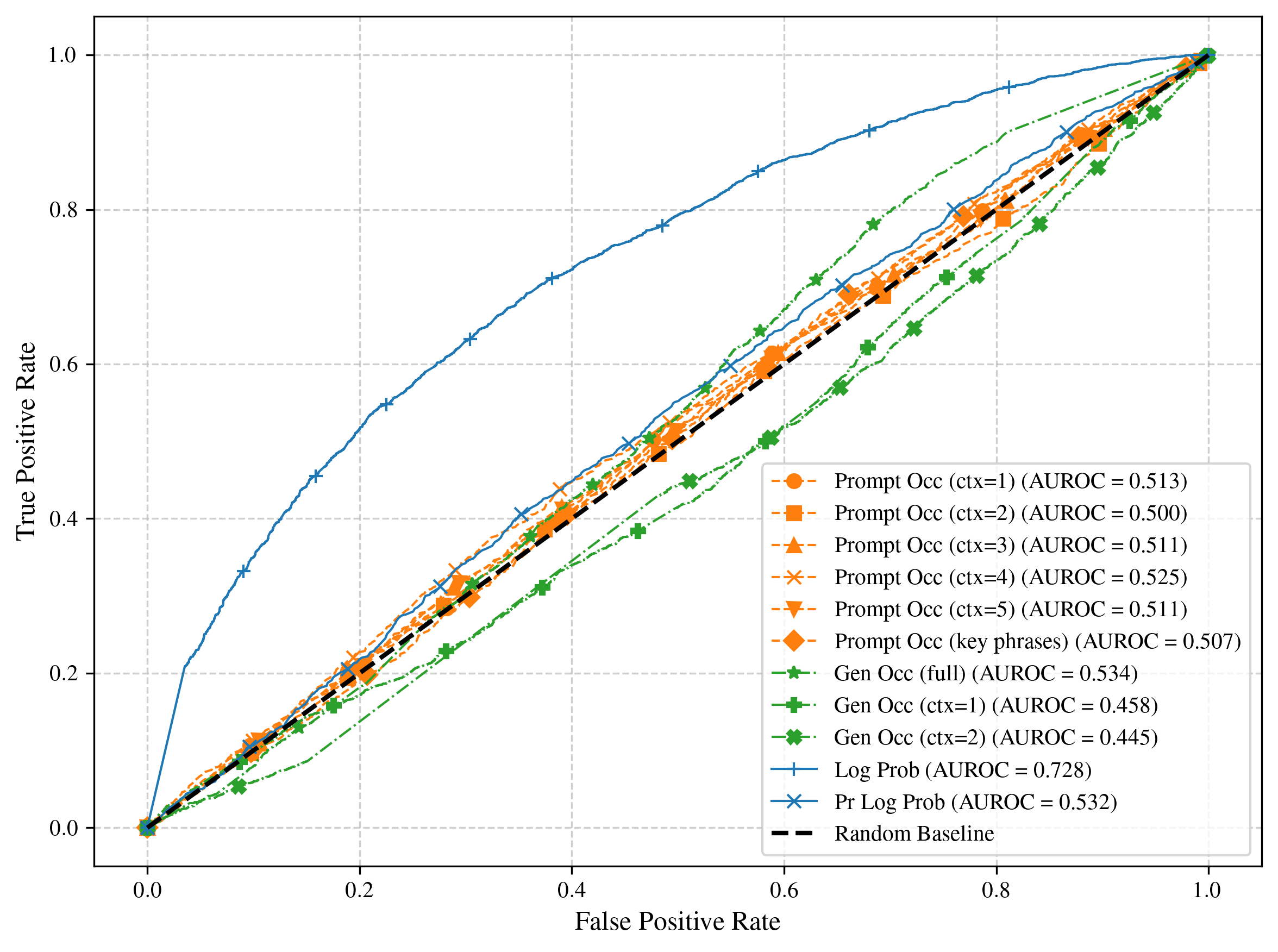}
    \caption{CoQA (Raw Frequency)}
\end{subfigure}
\hfill
\begin{subfigure}[t]{0.49\textwidth}
    \centering
    \includegraphics[width=\linewidth]{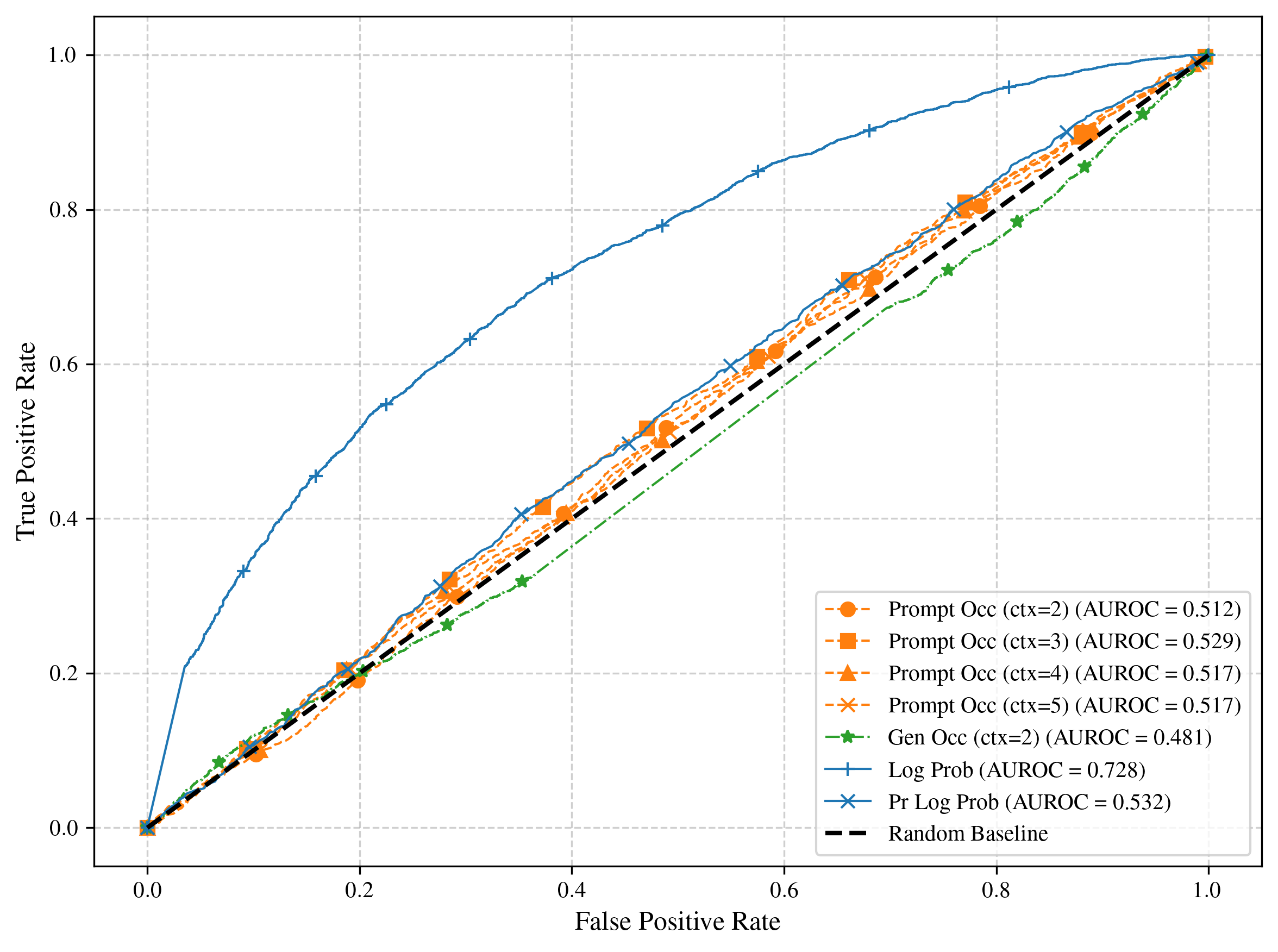}
    \caption{CoQA ($n$-gram Model)}
\end{subfigure}

\caption{AUROC curves comparing log-probabilities and occurrence-based features across datasets. Model: RedPajama-INCITE-3B; metric: RougeL}
\label{fig:auroc-all-3b}
\end{figure*}

Figure~\ref{fig:auroc-all-3b} shows the AUROC scores of raw frequency and $n$-gram features under the RedPajama-INCITE-3B model. We observe similar overall trends to the 7B model. Table~\ref{tab:acc_3b_rougeL_new} presents the corresponding classifier accuracy results.

\begin{table*}[!t]
\centering
\small
\resizebox{\textwidth}{!}{
\begin{tabular}{@{}lllc@{\hskip 6pt}cc|@{\hskip 6pt}cc@{}}
\toprule
\textbf{Feature Type} & \textbf{Dataset} & \textbf{Model} & \textbf{Tree Depth} & \textbf{Train Acc (Log)} & \textbf{Train Acc (Full)} & \textbf{Test Acc (Log)} & \textbf{Test Acc (Full)} \\
\midrule

\multirow{25}{*}{\textbf{Raw Freq}} & \multirow{8}{*}{TriviaQA} & \textit{Threshold} & --  & \textit{0.731 $\pm$ 0.004} & -- & \textit{0.724 $\pm$ 0.020} & -- \\
& & \multirow{6}{*}{Tree}
& 3  & 0.738 $\pm$ 0.003 & \textbf{0.746} $\pm$ 0.004 & 0.720 $\pm$ 0.016 & \textbf{0.749} $\pm$ 0.008 \\
& & & 5  & 0.747 $\pm$ 0.004 & \textbf{0.760} $\pm$ 0.005 & 0.717 $\pm$ 0.021 & \textbf{0.736} $\pm$ 0.013 \\
& & & 7  & 0.760 $\pm$ 0.004 & \textbf{0.788} $\pm$ 0.012 & 0.716 $\pm$ 0.015 & \textbf{0.732} $\pm$ 0.017 \\
& & & 9  & 0.780 $\pm$ 0.004 & \textbf{0.829} $\pm$ 0.011 & 0.705 $\pm$ 0.019 & \textbf{0.715} $\pm$ 0.015 \\
& & & 10 & 0.792 $\pm$ 0.004 & \textbf{0.855} $\pm$ 0.011 & 0.692 $\pm$ 0.021 & \textbf{0.709} $\pm$ 0.013 \\
& & & 20 & 0.923 $\pm$ 0.011 & \textbf{0.992} $\pm$ 0.004 & 0.664 $\pm$ 0.017 & \textbf{0.678} $\pm$ 0.027 \\
& & NN & -- & 0.735 $\pm$ 0.004 & \textbf{0.738} $\pm$ 0.005 & \textbf{0.730} $\pm$ 0.012 & 0.721 $\pm$ 0.013 \\

\addlinespace

& \multirow{8}{*}{NQ-Open} & \textit{Threshold} & --  & \textit{0.580 $\pm$ 0.004} & -- & \textit{0.585 $\pm$ 0.019} & -- \\
& & \multirow{6}{*}{Tree}
& 3  & 0.605 $\pm$ 0.011 & \textbf{0.628} $\pm$ 0.014 & \textbf{0.602} $\pm$ 0.032 & 0.551 $\pm$ 0.037 \\
& & & 5  & 0.622 $\pm$ 0.009 & \textbf{0.695} $\pm$ 0.004 & \textbf{0.586} $\pm$ 0.021 & 0.559 $\pm$ 0.037 \\
& & & 7  & 0.643 $\pm$ 0.016 & \textbf{0.770} $\pm$ 0.018 & \textbf{0.576} $\pm$ 0.027 & 0.548 $\pm$ 0.048 \\
& & & 9  & 0.668 $\pm$ 0.026 & \textbf{0.835} $\pm$ 0.024 & \textbf{0.585} $\pm$ 0.021 & 0.548 $\pm$ 0.044 \\
& & & 10 & 0.680 $\pm$ 0.026 & \textbf{0.868} $\pm$ 0.034 & \textbf{0.591} $\pm$ 0.033 & 0.541 $\pm$ 0.043 \\
& & & 20 & 0.793 $\pm$ 0.048 & \textbf{1.000} $\pm$ 0.000 & \textbf{0.548} $\pm$ 0.039 & 0.544 $\pm$ 0.042 \\
& & NN & -- & 0.586 $\pm$ 0.008 & \textbf{0.590} $\pm$ 0.018 & \textbf{0.590} $\pm$ 0.018 & 0.584 $\pm$ 0.027 \\

\addlinespace

& \multirow{8}{*}{CoQA} & \textit{Threshold} & --  & \textit{0.664 $\pm$ 0.003} & -- & \textit{0.665 $\pm$ 0.008} & -- \\
& & \multirow{6}{*}{Tree}
& 3  & 0.668 $\pm$ 0.002 & \textbf{0.668} $\pm$ 0.007 & 0.661 $\pm$ 0.009 & \textbf{0.664} $\pm$ 0.011 \\
& & & 5  & 0.674 $\pm$ 0.001 & \textbf{0.678} $\pm$ 0.004 & 0.664 $\pm$ 0.004 & \textbf{0.667} $\pm$ 0.017 \\
& & & 7  & 0.686 $\pm$ 0.003 & \textbf{0.700} $\pm$ 0.008 & \textbf{0.663} $\pm$ 0.006 & 0.651 $\pm$ 0.017 \\
& & & 9  & 0.701 $\pm$ 0.004 & \textbf{0.735} $\pm$ 0.013 & \textbf{0.655} $\pm$ 0.006 & 0.639 $\pm$ 0.012 \\
& & & 10 & 0.712 $\pm$ 0.004 & \textbf{0.756} $\pm$ 0.016 & \textbf{0.653} $\pm$ 0.008 & 0.633 $\pm$ 0.015 \\
& & & 20 & 0.827 $\pm$ 0.015 & \textbf{0.952} $\pm$ 0.013 & \textbf{0.627} $\pm$ 0.013 & 0.588 $\pm$ 0.012 \\
& & NN & -- & 0.666 $\pm$ 0.003 & \textbf{0.667} $\pm$ 0.004 & \textbf{0.666} $\pm$ 0.010 & 0.664 $\pm$ 0.010 \\

\midrule

\multirow{25}{*}{\textbf{N-gram}} & \multirow{8}{*}{TriviaQA} & \textit{Threshold} & --  & \textit{0.731 $\pm$ 0.004} & -- & \textit{0.724 $\pm$ 0.020} & -- \\
& & \multirow{6}{*}{Tree}
& 3  & 0.738 $\pm$ 0.003 & \textbf{0.741} $\pm$ 0.005 & 0.720 $\pm$ 0.016 & \textbf{0.724} $\pm$ 0.011 \\
& & & 5  & 0.747 $\pm$ 0.004 & \textbf{0.763} $\pm$ 0.004 & \textbf{0.717} $\pm$ 0.021 & 0.712 $\pm$ 0.018 \\
& & & 7  & 0.760 $\pm$ 0.004 & \textbf{0.793} $\pm$ 0.006 & \textbf{0.716} $\pm$ 0.015 & 0.701 $\pm$ 0.011 \\
& & & 9  & 0.780 $\pm$ 0.004 & \textbf{0.836} $\pm$ 0.009 & \textbf{0.705} $\pm$ 0.019 & 0.686 $\pm$ 0.015 \\
& & & 10 & 0.792 $\pm$ 0.004 & \textbf{0.862} $\pm$ 0.007 & \textbf{0.692} $\pm$ 0.021 & 0.683 $\pm$ 0.010 \\
& & & 20 & 0.923 $\pm$ 0.011 & \textbf{0.993} $\pm$ 0.005 & \textbf{0.664} $\pm$ 0.017 & 0.651 $\pm$ 0.014 \\
& & NN & -- & 0.735 $\pm$ 0.004 & \textbf{0.737} $\pm$ 0.003 & \textbf{0.730} $\pm$ 0.012 & 0.728 $\pm$ 0.012 \\

\addlinespace 

& \multirow{8}{*}{NQ-Open} & \textit{Threshold} & --  & \textit{0.580 $\pm$ 0.004} & -- & \textit{0.585 $\pm$ 0.019} & -- \\
& & \multirow{6}{*}{Tree}
& 3  & 0.605 $\pm$ 0.011 & \textbf{0.620} $\pm$ 0.005 & \textbf{0.602} $\pm$ 0.032 & 0.528 $\pm$ 0.030 \\
& & & 5  & 0.622 $\pm$ 0.009 & \textbf{0.690} $\pm$ 0.015 & 0.586 $\pm$ 0.021 & \textbf{0.609} $\pm$ 0.046 \\
& & & 7  & 0.643 $\pm$ 0.016 & \textbf{0.748} $\pm$ 0.019 & \textbf{0.576} $\pm$ 0.027 & 0.566 $\pm$ 0.036 \\
& & & 9  & 0.668 $\pm$ 0.026 & \textbf{0.803} $\pm$ 0.026 & \textbf{0.585} $\pm$ 0.021 & 0.583 $\pm$ 0.053 \\
& & & 10 & 0.680 $\pm$ 0.026 & \textbf{0.833} $\pm$ 0.028 & \textbf{0.591} $\pm$ 0.033 & 0.583 $\pm$ 0.041 \\
& & & 20 & 0.793 $\pm$ 0.048 & \textbf{0.986} $\pm$ 0.016 & 0.548 $\pm$ 0.039 & \textbf{0.555} $\pm$ 0.029 \\
& & NN & -- & 0.586 $\pm$ 0.008 & \textbf{0.601} $\pm$ 0.012 & \textbf{0.590} $\pm$ 0.018 & 0.576 $\pm$ 0.034 \\

\addlinespace

& \multirow{8}{*}{CoQA} & \textit{Threshold} & --  & \textit{0.664 $\pm$ 0.003} & -- & \textit{0.665 $\pm$ 0.008} & -- \\
& & \multirow{6}{*}{Tree}
& 3  & 0.668 $\pm$ 0.002 & \textbf{0.671} $\pm$ 0.006 & 0.661 $\pm$ 0.009 & \textbf{0.666} $\pm$ 0.012 \\
& & & 5  & 0.674 $\pm$ 0.001 & \textbf{0.685} $\pm$ 0.005 & 0.664 $\pm$ 0.004 & \textbf{0.673} $\pm$ 0.010 \\
& & & 7  & 0.686 $\pm$ 0.003 & \textbf{0.711} $\pm$ 0.006 & \textbf{0.663} $\pm$ 0.006 & 0.653 $\pm$ 0.006 \\
& & & 9  & 0.701 $\pm$ 0.004 & \textbf{0.750} $\pm$ 0.010 & \textbf{0.655} $\pm$ 0.006 & 0.642 $\pm$ 0.016 \\
& & & 10 & 0.712 $\pm$ 0.004 & \textbf{0.774} $\pm$ 0.012 & \textbf{0.653} $\pm$ 0.008 & 0.630 $\pm$ 0.020 \\
& & & 20 & 0.827 $\pm$ 0.015 & \textbf{0.962} $\pm$ 0.018 & \textbf{0.627} $\pm$ 0.013 & 0.598 $\pm$ 0.005 \\
& & NN & -- & 0.666 $\pm$ 0.003 & \textbf{0.672} $\pm$ 0.005 & 0.666 $\pm$ 0.010 & \textbf{0.673} $\pm$ 0.008 \\

\bottomrule
\end{tabular}
}
\caption{Accuracy across tree depths, datasets, and feature types. Bold indicates higher accuracy between log-only and full features. Model: RedPajama-INCITE-3B; metric: ROUGE-L.}
\label{tab:acc_3b_rougeL_new}
\end{table*}

\subsection{Data Sparsity Analysis}
\label{sec: appendix-sparsity}
Table~\ref{tab:data sparsity} reports the proportion of $n$-grams missing from the RedPajama training corpus across different test datasets, along with representative examples. Overall, a substantial proportion of 4-grams and longer sequences are absent verbatim from the training data, particularly in open-domain questions.

\begin{table*}[t!]
\centering
\small
\resizebox{\textwidth}{!}{
\begin{tabular}{@{}llc@{\hskip 10pt}c@{\hskip 10pt}p{0.55\textwidth}@{}}
\toprule
\textbf{Dataset} & \textbf{N-gram} & \textbf{\% Zeros} & \textbf{Examples with zero occurrence in training data} \\
\midrule
\multirow{6}{*}{TriviaQA}
& 1           & 0.00    & -- \\
& 2           & 0.03    & \texttt{mental originate}\quad\textbf{|}\quad\texttt{aston ban}\quad\textbf{|}\quad\texttt{ovsk refer} \\
& 3           & 3.33    & \texttt{keyboard the \%}\quad\textbf{|}\quad\texttt{iva' released}\quad\textbf{|}\quad\texttt{56 balls?} \\
& 4           & 14.97   & \texttt{Jersey Boys""}\quad\textbf{|}\quad\texttt{ove are types of}\quad\textbf{|}\quad\texttt{which famous product?} \\
& 5           & 31.99   & \texttt{what is classified using the}\quad\textbf{|}\quad\texttt{Who (at 2008)}\quad\textbf{|}\quad\texttt{"In Texas, what} \\
& key phrases & 41.30   & \texttt{there will be blood, film}\quad\textbf{|}\quad\texttt{actor, second Dr Who}\quad\textbf{|}\quad\texttt{December 2, 1805, battle} \\
\addlinespace
\addlinespace
\multirow{6}{*}{NQ-Open}
& 1           & 0.00    & -- \\
& 2           & 0.06    & \texttt{slew win}\quad\textbf{|}\quad\texttt{today gest}\quad\textbf{|}\quad\texttt{gium invade} \\
& 3           & 4.79    & \texttt{when does bill}\quad\textbf{|}\quad\texttt{s eg wells}\quad\textbf{|}\quad\texttt{plays dorian} \\
& 4           & 21.09   & \texttt{build tables that identify}\quad\textbf{|}\quad\texttt{ondon finished being built}\quad\textbf{|}\quad\texttt{ sings oh what a} \\
& 5           & 44.44   & \texttt{boundary was the mexico}\quad\textbf{|}\quad\texttt{who played zoe h}\quad\textbf{|}\quad\texttt{ does the paraguay} \\
& key phrases & 55.48   & \texttt{completion of tower of london}\quad\textbf{|}\quad\texttt{ubuntu project founder}\quad\textbf{|}\quad\texttt{macos operating system} \\

\addlinespace
\addlinespace
\multirow{6}{*}{CoQA}
& 1           & 0.00    & -- \\
& 2           & 0.02    & \texttt{TribeII}\quad\textbf{|}\quad\texttt{ Collect Doyle}\quad\textbf{|}\quad\texttt{ compensatory innocence} \\
& 3           & 1.56    & \texttt{golden dog Kay}\quad\textbf{|}\quad\texttt{upon, Johann}\quad\textbf{|}\quad\texttt{ Rex ran home} \\
& 4           & 9.40    & \texttt{of prosecutors argument that}\quad\textbf{|}\quad\texttt{ away shooting; for}\quad\textbf{|}\quad\texttt{ to learn French signs} \\
& 5           & 24.80   & \texttt{asked king to send him}\quad\textbf{|}\quad\texttt{ man on that houseboat}\quad\textbf{|}\quad\texttt{ Charity found Arthur would} \\
& key phrases & 39.37   & \texttt{sheriff norman chaffins}\quad\textbf{|}\quad\texttt{ nokobee woods}\quad\textbf{|}\quad\texttt{ new york fashion internships} \\
\bottomrule
\end{tabular}
}
\caption{Percentage of zero-occurrence n-grams and key phrases in the training corpus}
\label{tab:data sparsity}
\end{table*}

\end{document}